\documentclass[preprint,12pt]{elsarticle}
\usepackage[margin=2.5cm]{geometry}
\usepackage{xcolor}
\usepackage{booktabs}
\usepackage{setspace}
\usepackage{placeins}
\usepackage{booktabs}
\usepackage{adjustbox}
\usepackage{tabularx}
\usepackage{hyperref}
\usepackage{pdflscape}
\usepackage{caption}
\usepackage{longtable}
\usepackage{float}
\usepackage{soul}
\usepackage{setspace}
\usepackage{caption}
\usepackage{hypcap}
\usepackage{array}
\usepackage[utf8]{inputenc}
\usepackage[T1]{fontenc}
\DeclareUnicodeCharacter{03BC}{\ensuremath{\mu}}
\usepackage{gensymb}

\hypersetup{
    colorlinks=true,
    linkcolor=blue,
    filecolor=magenta,      
    urlcolor=cyan
}

\usepackage{amssymb}
\usepackage{amsmath}
\usepackage{textcomp}

\begin{document}

\begin{frontmatter}


\title{AMGPT: a Large Language Model for Contextual Querying in Additive Manufacturing}




\author[inst1]{Achuth Chandrasekhar}
\ead{achuthc@andrew.cmu.edu}

\affiliation[inst1]{organization={Materials Science and Engineering},
            addressline={Carnegie Mellon University}, 
            city={Pittsburgh},
            postcode={15213}, 
            state={PA},
            country={USA}}

\author[inst2]{Jonathan Chan}
\ead{jchan5@andrew.cmu.edu}
\author[inst2]{Francis Ogoke}
\ead{oogoke@andrew.cmu.edu}
\author[inst2]{Olabode Ajenifujah}
\ead{oajenifu@andrew.cmu.edu}
\author[inst2,inst3,inst4,inst5]{Amir Barati Farimani\corref{cor1}}
\ead{barati@cmu.edu}
\cortext[cor1]{Corresponding author}

\affiliation[inst2]{organization={Mechanical Engineering},
            addressline={Carnegie Mellon University}, 
            city={Pittsburgh},
            postcode={15213}, 
            state={PA},
            country={USA}}

\affiliation[inst3]{organization={Biomedical Engineering},
            addressline={Carnegie Mellon University}, 
            city={Pittsburgh},
            postcode={15213}, 
            state={PA},
            country={USA}}
\affiliation[inst4]{organization={Chemical Engineering},
            addressline={Carnegie Mellon University}, 
            city={Pittsburgh},
            postcode={15213}, 
            state={PA},
            country={USA}}

\affiliation[inst5]{organization={Machine Learning Department},
            addressline={Carnegie Mellon University}, 
            city={Pittsburgh},
            postcode={15213}, 
            state={PA},
            country={USA}}


\begin{abstract}
Generalized large language models (LLMs) such as GPT-4 may not provide specific answers to queries formulated by materials science researchers. These models may produce a high-level outline but lack the capacity to return detailed instructions on manufacturing and material properties of novel alloys. Enhancing a smaller model with specialized domain knowledge may provide an advantage over large language models which cannot be retrained quickly enough to keep up with the rapid pace of research in metal additive manufacturing (AM). We introduce "AMGPT," a specialized LLM text generator designed for metal AM queries. The goal of AMGPT is to assist researchers and users in navigating the extensive corpus of literature in AM. Instead of training from scratch, we employ a pre-trained Llama2-7B model from Hugging Face in a Retrieval-Augmented Generation (RAG) setup, utilizing it to dynamically incorporate information from $\sim$50 AM papers and textbooks in PDF format. Mathpix is used to convert these PDF documents into TeX format, facilitating their integration into the RAG pipeline managed by LlamaIndex. Expert evaluations of this project highlight that specific embeddings from the RAG setup accelerate response times and maintain coherence in the generated text.

\end{abstract}



\end{frontmatter}



\section{Introduction}
\label{sec:introduction}

Since the introduction of the Transformer architecture in 2017 \cite{aiayn}, large language models (LLMs) have transformed the field of word sequence prediction. This architecture effectively addresses the vanishing gradient problem, which commonly arises in traditional recurrent neural networks (RNNs) during backpropagation, marking a significant advancement in deep learning technologies. Now, LLMs can generate long chunks of text spanning paragraphs instead of sentences. They accomplish this by processing millions of web-scraped pages off the internet, converting them to text, and predicting the next most likely sequence. When trained on a large corpus of human-written text, the language model picks up patterns from a probabilistic approach, rather than a rational, principle-driven one. Despite their lack of logic, LLMs have seen great success in various industries as models for scientific discovery and inquiry \cite{bran2023chemcrow}, \cite{mccabe2023multiple}, \cite{lanusse2023astroclip}. 

Popularized by the public launch of GPT-3.5 in 2022 \cite{gpt3}, LLMs belong to the family of foundation models, which enable generative AI applications. They can do question answering, known as chat, or step-by-step instruction generation, known as instruct. LLMs can be classified into three main categories. Encoder-only models are designed to classify sequences of text, which is primarily associated with natural language understanding tasks. Decoder-only models focus on generating text, which aligns with natural language generation. Encoder-decoder models perform sequence-to-sequence conversions, effectively handling tasks that involve transforming one form of textual input into another. The advantage of incorporating an encoder in the model is context, where prior tokens in the prompt are directly used in the response \cite{zhu2024large}. These qualities make LLMs versatile and teachable. 

LLMs leverage several key components: prompts, training data, model architectures, embeddings, vector databases, and similarity search algorithms. Prompts provide the necessary input for querying and generating text in decoder models. Training data offer LLMs a vast corpus of references, enabling them to learn and predict probable word connections. The model architectures, such as GPT \cite{Radford2018ImprovingLU}, T5 \cite{Raffel2019ExploringTL}, BART \cite{Lewis2019BARTDS}, Mixtral \cite{Jiang2024MixtralOE}, Llama \cite{Touvron2023LLaMAOA}, and BERT \cite{devlin-etal-2019-bert}, enable LLMs to learn complex relationships and generate coherent text. Embeddings transform words into high-dimensional numerical vectors, facilitating computational operations. Vector databases store these embeddings, while similarity search algorithms retrieve similar sequences by parsing these vector stores.

By fine tuning LLMs on data containing domain knowledge, they can be transformed into experts in specific subjects. These models particularly excel in fields where the volume of literature is too vast for humans to thoroughly review and organized understanding is required to find relevant information specific to the semantics of a query \cite{Agarwal2024LitLLMAT}. For example, models have been trained on medical journals and clinical trials to provide accurate opinions on patient readmission risks and medical prognosis one year after treatment. BERT-based models that have achieved state-of-the-art performance in various Natural Language Processing tasks include BIOBERT \cite{Lee2019BioBERTAP}, which excels in biomedical text mining, CLINICALBERT \cite{Huang2019ClinicalBERTMC}, which is optimized for clinical text interpretation, and SciBERT \cite{Beltagy2019SciBERTAP}, which targets scientific publications. 

Expanding the applications of language models beyond natural language processing, several domain-specific models have emerged to address challenges in various scientific and engineering applications. MechGPT \cite{Buehler2024GenerativeRO} is a pioneering language model developed to integrate knowledge across various scales and modalities in the fields of mechanics and materials science. Similar to MechGPT, other domain-specific language models such as ChemBERTa \cite{Chithrananda2020ChemBERTaLS} for chemical compound predictions, Materials Project Informer (MPI) for materials discovery, GeoBERT \cite{Gao2022GeoBERTPG} for geological data analysis, CatBERTa for catalyst adsorption energy prediction \cite{Ock2023CatalystPP} and BioBART \cite{yuan2022biobart} for biomedical text analysis have also demonstrated the utility of fine-tuned language models in addressing complex interdisciplinary challenges in science and engineering.

While fine-tuning LLMs reduces the frequency of hallucinations—instances where the model generates plausible-sounding but incorrect or nonsensical information—and increases task-specific performance, it does not completely eliminate them. Other key challenges include the need for large, specialized training datasets and confinement to a strict question-response format. Technical constraints also exist, such as limited context token capacity and high computational demands, with training and operation requiring several petaflops of processing power and extensive GPU memory. Similar to how large datasets filled with metrics about patient genetics, blood composition, and lifestyle can be used to fine tune LLMs such as ClinicalT5 \cite{lu-etal-2022-clinicalt5}, papers on metal additive manufacturing can be used to fine tune a pre-trained LLM. However, additive manufacturing presents unique challenges, such as the need to understand complex material behaviors, process parameters, and the interactions between different stages of production. In this field, the development of large language models  that reliably and accurately reference knowledge from journal papers remains relatively underexplored. While Authentise\footnote{\scriptsize{\url{https://www.authentise.com/post/authentise-brings-chatgpt-capabilities-to-additive-manufacturing}}} has proposed 3DGPT, which builds on a pre-trained ChatGPT model to generate reference citations on inference, it is proprietary and does not contain specific considerations to reduce hallucinatory responses from the model.

 For an LLM to be considered a domain knowledge expert, it should be capable of directly extracting relevant excerpts from primary sources. For example, users can ask questions like "How do I reduce the likelihood of defects when working with Aluminum alloys in laser powder bed fusion?" and LLMs could potentially offer them intelligent insights comparable to those of human subject experts. 

\parindent=.5cm

Retrieval-Augmented Generation (RAG), introduced by Lewis et al. \cite{lewis_rag}, represents a pivotal advancement in enhancing the capabilities of LLMs by integrating dynamic, external knowledge bases. This synthesis of inherent model knowledge with vast, updated external data sources aims to address the challenges LLMs face, such as hallucinations, and to improve transparency in reasoning processes. RAG introduces a novel method for customizing LLMs to specific tasks by retrieving specific text data relevant to user queries, thus augmenting the base functionality of LLMs with specialized knowledge. This process boosts LLM performance by enabling access to insightful data from the training set itself, as shown by Khandelwal et al. \cite{Khandelwal2019GeneralizationTM}.

In the field of metal additive manufacturing, LLMs that integrate RAG with comprehensive procedural guidelines (LLM-RAG) are likely to provide more reliable outputs. This integration allows the language models to access and utilize vast repositories of domain-specific knowledge and procedural data, ensuring that their responses are both contextually accurate and aligned with established best practices. By leveraging RAG, LLMs can retrieve pertinent information on demand, enhancing their ability to provide detailed, precise, and up-to-date guidance tailored to the specific needs of additive manufacturing processes. This leads to improved decision-making, reduced errors, and increased efficiency in the manufacturing workflow. Presently, two leading open-source platforms support RAG: LangChain\footnote{Langchain is available at \url{https://github.com/langchain-ai/langchain}} and LlamaIndex\footnote{LlamaIndex is available at \url{https://github.com/jerryjliu/llama\_index}}. These two platforms have made RAG very modular and customizable over a wide range of technical applications.

These platforms form the basis upon which we propose AMGPT, an open-source, medium-sized language model consisting of 7 billion parameters. AMGPT utilizes RAG in conjunction with a chat memory approach to produce cohesive, evidence-based responses to queries, thereby enhancing the decision-making process in additive manufacturing. Our contributions, detailed below, reflect the practical implementation and customization of AMGPT to address the unique challenges in this field:
\parindent=.7cm

\begin{enumerate}
    \item Characterize the effectiveness of RAG methods on pre-trained models.
    \item Create a subject matter expert LLM using a corpus of additive manufacturing papers.
    \item Develop an end-to-end pipeline for deploying a chatbot, from acquiring pre-trained models from Hugging Face to implementing a user interface powered by Streamlit.
\end{enumerate}

\section{Related Works}
\label{sec:relatedworks}
\subsection{Deep Learning for Additive Manufacturing}

As the complexity and scale of additive manufacturing (AM) continue to grow, deep learning has emerged as a more potent solution for addressing AM challenges than traditional machine learning. Deep learning excels because it can handle vast and intricate datasets, identify sophisticated nonlinear patterns, and provide swift predictions.

Part design can benefit from a data-driven topology optimization approach that integrates geometrical manufacturing constraints early in the design phase. This method, presented by Almasri et al. \cite{almasri2023data}, leverages the Deep Learning Additive Manufacturing driven Topology Optimization (DL-AM-TO) model to enhance the optimization process, demonstrating improved efficiency and feasibility in creating manufacturable parts by addressing constraints at the initial design steps.

Post-printing quality control in additive manufacturing is enhanced by ThermoPore, a deep learning model that predicts part porosity from thermal images using in-situ monitoring data from Laser Powder Bed Fusion (LPBF) processes. Pak et al. \cite{pak2024thermopore} introduced ThermoPore, demonstrating its potential to improve defect prediction and reduce post-inspection needs. Similarly, Estalaki et al. \cite{estalaki2022predicting} developed machine learning models to predict microporosity in LPBF stainless steel materials using in-situ thermographic data. 

Convolutional neural networks (CNNs) have been utilized for in-situ monitoring of AM processes to accurately identify defects and anomalies during the printing process, enabling immediate corrective actions. This approach significantly enhances the reliability of AM systems and reduces the need for post-process inspection, as demonstrated by Scime et al. \cite{Scime2018} and Pandiyan et al. \cite{pandiyan2022situ}.

In conclusion, deep learning offers substantial benefits for additive manufacturing, from real-time defect detection to process optimization. While the applications mentioned target specific issues within the additive manufacturing process pipeline, AMGPT represents a broader application of language models in AM. As a large language model with retrieval-augmented generation (LLM-RAG) capabilities, AMGPT can provide contextual assistance, streamline decision-making processes, and enhance knowledge management across various AM tasks. This flexibility makes it a valuable tool in both specific and general contexts within the field.

\subsection{Medium Language Models}
A study by OpenAI \cite{NEURIPS2022_b1efde53} shows that adding more parameters does not directly boost the performance of response generation. Rather, larger models tend to hallucinate. The main benefit of additional parameters is the precision of the output text, where a larger tokenizer leads to a more comprehensive dictionary. OpenAI found that inferences may not necessarily be more accurate with increasing numbers of parameters, as evaluated by humans assigning preferences for one LLM’s response over the other. Other studies have found success in bootstrapping medium-sized language models on the order of $10^{9}$ parameters \cite{mukherjee2024polaris} or utilizing a mixture of experts of medium size by initializing more feedforward blocks to expand token windows \cite{Jiang2024MixtralOE}. For our purposes, we select models with 1-10B parameters, such as Llama and Mixtral, with open source add-ins for versatility.

\subsection{Quantization}
Minimizing the model size during training and inference while maintaining LLM performance is an implementation objective called quantization. Loading models for inference requires a large amount of GPU memory. The rule of thumb is every billion parameters requires 3 GB of GDDR6 VRAM for the default precision of parameter values \cite{huggingface2023optimize}, \cite{huggingface2024llama}, \cite{huggingface2024optimize}. Floating point numbers contain 16 bits, which supplies enough distinct values for computation and discerning meanings behind encoded text information. Quantization recasts these model weights into lower precision data types such as FP8 or FP4, which slightly reduces performance at the benefit of requiring half or a quarter of the memory requirement. Success has been shown in LLM-FP4 \cite{liu-etal-2023-llm}, where floating point (FP) quantization offers versatility and can approximate normal distributions better than integer quantization like int8. Since weights should be diversely initialized when training as in Xavier initialization\cite{glorot10a}, FP quantization helps LLMs learn. Quantization need not be performed during training. Post-training quantization (PQR) reduces computational overhead and memory consumption in LLMs \cite{gong2024makes}\cite{li2024evaluating}. Our system consists of two NVIDIA A6000 GPUs, each having 48 GB of memory, sufficient for any model below 32B parameters with FP16 data. The embedding model, active tokens from the database, any augmentation with bootstrapped models, and computation from a running script will occupy additional memory, making the realistic limit 29B parameters. We opt for high performing 7B language models, loading them in quantized as FP16 and FP8 and did not use mixed integer quantization for our application. With memory demands resolved, we can interact with the LLM in downstream natural language tasks. 

\subsection{Fine-tuning}
Fine tuning LLMs for domain expertise requires a custom data set and structured data within the field of knowledge. The author of MechGPT \cite{mechgpt2024buehler} uses a 70B Llama2 Chat model to distill scientific question-answer pairs from a textbook. The book is divided into 430 chunks, each having around 500 words, then a two-step process is used to first generate hypothesis-driven or characterization questions based on each chunk and secondly concisely summarize it to form the answers. This scalable system can be adapted to any text source. 

Fine tuning LLMs also requires substantial compute resources. Within the realm of medium language models that we aim to use, for example, the Llama2 7B model took 184,320 GPU hours on a A100 GPU with 80 GB capacity to train \cite{touvron2023llama}. Due to the computational expense of training a model from scratch, fine tuning the LLM for more epochs after the model is initially trained, which updates all its weights, is unfeasible. Consequently, more data efficient methods have been developed. Low rank adaptation (LoRA) interweaves trainable rank decomposition matrices into each Transformer layer \cite{hu2021lora}. Instead of retraining all the parameters, they are frozen and the architecture is augmented with a few more layers, which have four orders of magnitude less trainable weights and require a third of the memory requirement. QLoRA offers an even more streamlined approach, innovating a new data type called 4-bit NormalFloat (NF4) that optimizes the representation for normally distributed weights \cite{dettmers2023qlora}. These rank adaptation methods for the matrices that contain model parameters enhance performance on benchmarks and domain-specific tasks. Moreover, pre-trained LLMs exhibit a capacity for few-shot generalization on downstream natural language tasks without the need for extensive fine tuning \cite{gpt3}.

\subsection{Retrieval-Augmented Generation}
Retrieval-Augmented Generation (RAG) circumvents the need to retrain the parameter models or modify the model architecture in order to generate responses based on concrete evidence. By referring back to the original source material in the response, RAG reduces hallucinations,   as shown in a study where RAG boosted the accuracy of GPT4.0 from 80.1\% to 91.4\% in generating preoperative instructions as evaluated by an expert panel and against healthcare guidelines \cite{ke2024development}. 

Improving upon foundational models is extremely costly as the main role during training is to embed general natural language understanding through semantics and grammar. Tuning parametric memory requires intense computational resources and time, so utilizing nonparametric memory through a RAG model boosts accuracy in specific domains at substantially lower expense. The generator and retriever models, once pre-trained, can be bootstrapped to a base LLM to produce more factual, precise, and varied responses than non-bootstrapped seq2seq models \cite{lewis2021retrievalaugmented}. Such strategies with reasoning-and-acting (Re-Act) agents \cite{Yao2023LANGUAGEM} have seen success in retrieving material properties \cite{chiang2024llamp}, interfacing with APIs for research paper databases, such as Arxiv and Pubmed, for information on any published topic \cite{lála2023paperqa} and designing mechanical structures \cite{jadhav2024large}. For our application, RAG is sufficient to parse documents within a specific domain for a pre-trained medium language model to produce well-informed responses. 

Combining RAG with prompt engineering strategies use LLMs to facilitate the iterative process of query and answer generation. Inspired by the paper that shows that LLMs are zero-shot reasoners, meaning that no prior examples or requisite knowledge is supplied with the query, appending “Let’s think step by step” substantially boosts LLM performance on generating novel and accurate responses in a technique known as chain of thought (CoT) prompting \cite{wei2023chainofthought}. Augmenting CoT with RAG produces Retrieval-Augmented Thought (RAT), which calls the RAG model every step of thinking, allowing the LLM to dynamically expand the number of relevant tokens it can access.

\section{Methods}
\label{sec:methods}

\subsection{Model}

The RAG system described herein employs the LLaMA2-7b \cite{touvron2023llama} model due to its advanced capabilities in understanding and generating natural language. LLaMA2, known for its balance between computational efficiency and performance across diverse NLP tasks, serves as the foundation for our generation process. The model is integrated using the Hugging Face Transformers library, which facilitates easy access to pre-trained models and the implementation of custom workflows.

\subsection{Retrieval Mechanism}
\begin{figure}[hbt!]

\includegraphics[width=1\linewidth]{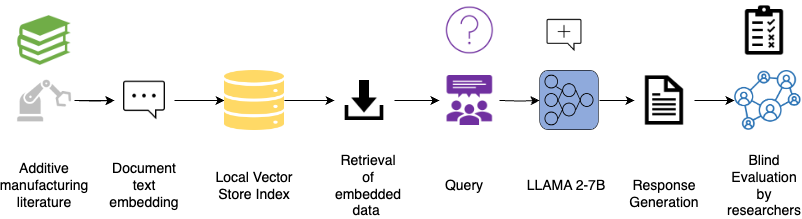}
\caption{A flowchart depicting the process of querying a database using an embedding model to provide context to a Large Language Model (LLM), which in turn generates an answer.
}
\label{fig:transformer_encoders}
\end{figure}

Our RAG system implements a dual-encoder retrieval mechanism, comprising a query encoder and a document encoder. Both encoders are fine-tuned versions of transformer-based models, optimized to encode text inputs into high-dimensional vectors. The query encoder transforms the input prompt into a query vector, while the document encoder maps documents from a pre-established corpus into corresponding vectors in the same semantic space. This setup allows for the efficient retrieval of documents most relevant to the input prompt, based on cosine similarity \cite{mikolov2013efficient} measures between the query vector and document vectors.

\subsection{Integration with Hugging Face}
The implementation leverages the Hugging Face ecosystem, particularly the transformers and datasets libraries, to streamline model training, fine-tuning, and deployment processes. The transformers library provides a comprehensive suite of tools for working with LLaMA2, including pre-trained model weights and tokenizer configurations. The datasets library supports the management of the retrieval corpus, enabling efficient indexing and querying operations crucial for the RAG's real-time information retrieval capabilities. 

\newpage

\subsection{Embedding model}
\begin{figure}[hbt!]
\includegraphics[width=1\linewidth]{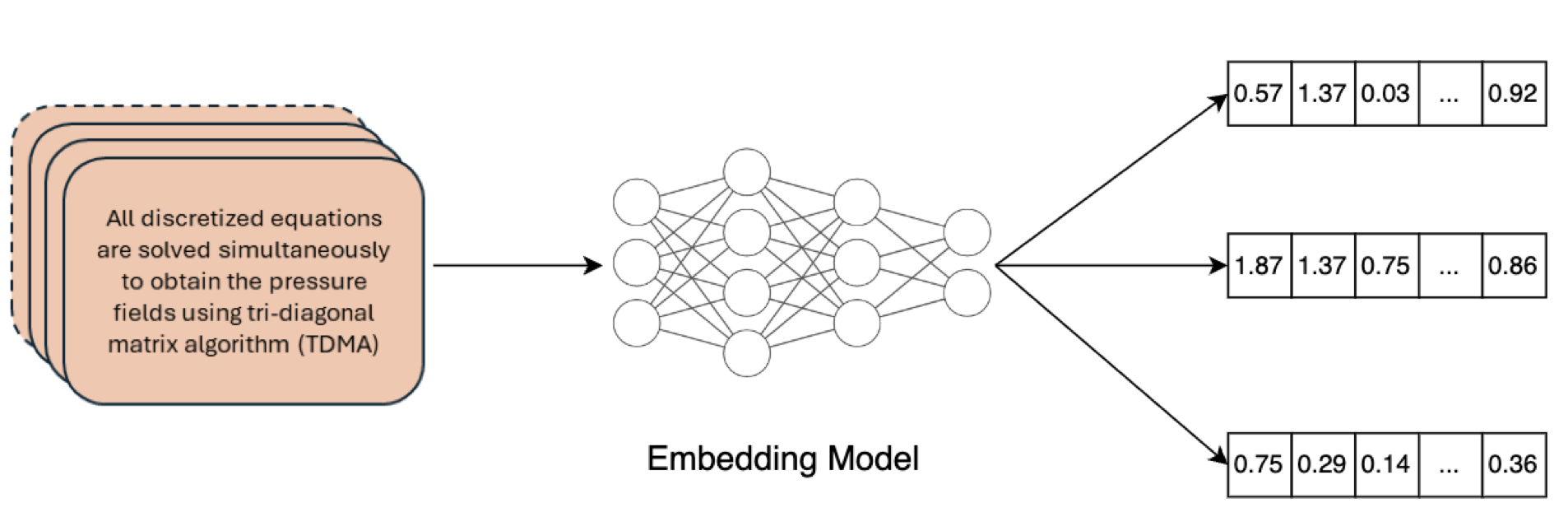}
\caption{Text embeddings are a fundamental technique in natural language processing (NLP) that translate words, sentences, or even entire documents into numerical vectors of real numbers. This process effectively converts discrete textual information into a form that can be understood and processed by machine learning algorithms. The figure shown above illustrates this process of abstraction by means of an embedding model. The primary aim of text embeddings is to encapsulate the semantic meaning of text, reflecting nuances of language such as context, tone, and similarity between words or phrases. 
}
\label{fig:emb_model}
\end{figure}

In order to encode natural language into numeric input, an embedding model transforms strings to a high dimensional vector space as shown in Figure \ref{fig:emb_model}. Generally, embeddings can convert any data space into a vectorized representation of each element, enabling multi-modal applications with data types varying from images to audio. For this paper, massive text embeddings are the primary focus.

As discussed in the Related Works section, though the language model itself can be reparameterized into a lower precision float or integer, the embedding model is not quantized.

Semantic search is different from standard keyword search by focusing on the capture of the intention of the user's query \cite{weisemantic}. At the heart of semantic search lies embedding models. SentenceTransformers \cite{reimers-gurevych-2019-sentence} is a python library that contains a vast corpus of BERT-style transformer models \cite{devlin-etal-2019-bert} that are fine-tuned for semantic search. The particular embedding model used here is "sentence-transformers/all-mpnet-base-v2", as illustrated in Figure \ref{fig:stage1}.

\subsection{Mechanisms and Concepts of Text Embeddings}

\textbf{Contextual Embeddings:} Models like ELMo (Embeddings from Language Models) \cite{elmo} and BERT (Bidirectional Encoder Representations from Transformers) generate representations that consider the entire sentence or passage, capturing the meaning of a word based on its context.

\textbf{Vector Space Models and Dimensionality Reduction:} Text embeddings map words or text to vectors in a high-dimensional space, where geometric relationships reflect semantic relationships. Embedding techniques often reduce dimensionality, compressing high-dimensional vectors into lower-dimensional spaces while preserving key relationships. This aids in capturing semantic and syntactic similarities and allows for efficient computation.

\textbf{Neural Networks:} Modern embeddings rely heavily on neural networks, from shallow models for word embeddings to complex transformer architectures in models like BERT and GPT for deeper context understanding.

\subsection{Devices and Codebase}

We employed the LLAMA2-7b chat model and the  sentence-transformers/all-mpnet-base-v2 embedding model, operating on a local NVIDIA RTX A6000 GPU. For public benefit and further research the code is available at the following link: \url{https://github.com/BaratiLab/LLM_RAG}. 

\vspace{15pt}


\begin{singlespace}
\textbf{\captionof{table}{Components and Tools for RAG}\label{table:retrieval_example}}
\begin{center}
\begin{tabularx}{\linewidth}{|>{\raggedright\arraybackslash}b{1.5cm}|>{\raggedright\arraybackslash}b{3.4cm}|>{\raggedright\arraybackslash}b{3.4cm}|>{\raggedright\arraybackslash}b{3.0cm}|>{\raggedright\arraybackslash}p{3.1cm}|}
\hline
& \textnormal{Preprocessing} & \textnormal{Embedding Model} & \textnormal{Vector Storage} & \textnormal{Retrieval Agent} \\
\hline
\textnormal{Function} & \textnormal{Helps convert PDF documents into TeX format} & \textnormal{Helps convert chunks into their vectors} & \textnormal{A storing space for all converted vectors} & \textnormal{A proxy to retrieve the most relevant documents} \\
\hline
\textnormal{Common Tools} & \textnormal{Mathpix: PDF to TeX} & \textnormal{Hugging Face (Local): sentence-transformers/all-mpnet-base-v2} & \textnormal{Local: Simple VectorStoreIndex, LlamaIndex} & \textnormal{LlamaIndex: Query Engine, Chat Engine} \\
\hline
\end{tabularx}
\end{center}
\end{singlespace}

\newpage

\subsection{Operational Workflow}
\begin{figure}[hbt!]
\centering
\includegraphics[width=1\linewidth]{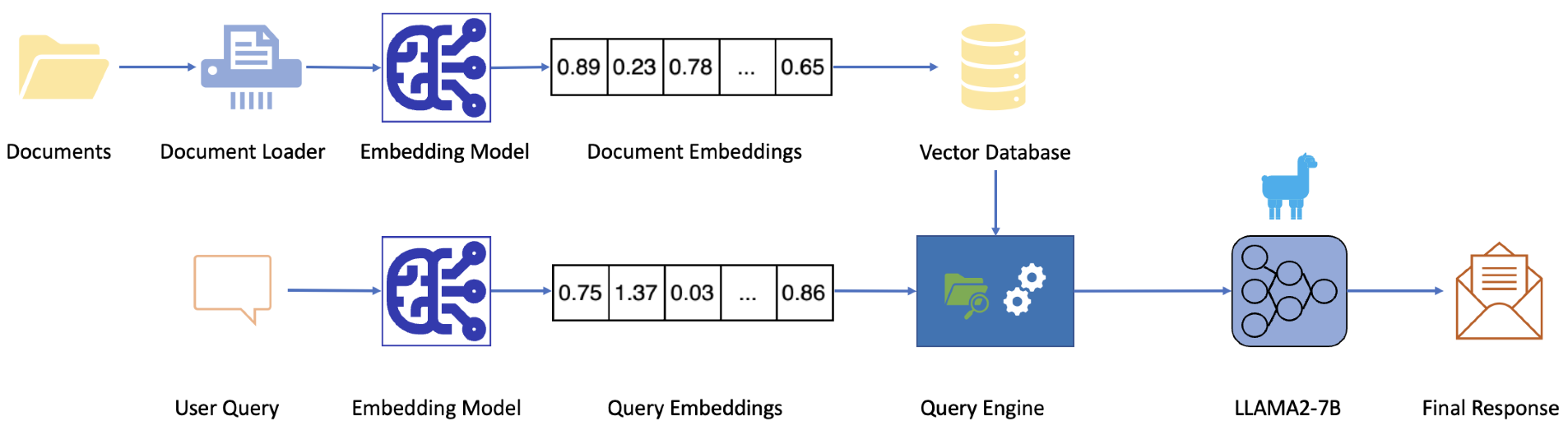}
\caption{Illustration of a Retrieval-Augmented Generation (RAG) workflow. Documents are loaded and processed into chunks, which are then embedded using an embedding model, creating vectors stored in a database. The query engine utilizes these vectors to match user queries against document chunks, and retrieves the most relevant content. Finally, the retrieved information is enhanced by the LLaMA2-7B language model to generate comprehensive and contextual responses.}
\label{fig:stage1}
\end{figure}

RAG is conducted with LangChain and LlamaIndex as shown in Figure \ref{fig:stage1} in order to facilitate the extraction of relevant mathematical expressions from the corpus. The selected PDF documents are converted into LaTeX form using Mathpix\footnote{\url{https://mathpix.com/}}, an API for document processing and conversion. The embedding model is called to generate the document embedding vectors, which are then stored in the VectorStoreIndex, a vector database provided by LlamaIndex. This database is subsequently used as the retrieval corpus.

Upon receiving an input prompt, the system first processes it through the embedding model, which acts as a query encoder to generate a corresponding query vector. This vector is then used to identify the most relevant documents from the retrieval corpus, based on similarity scoring. The selected documents, along with the original prompt, are concatenated and fed into the LLaMA2 model for response generation \cite{li2022survey}. 

\newpage

\subsection{Chat Interface}

\begin{figure}[hbt!]
\centering
\includegraphics[width=0.68\linewidth]{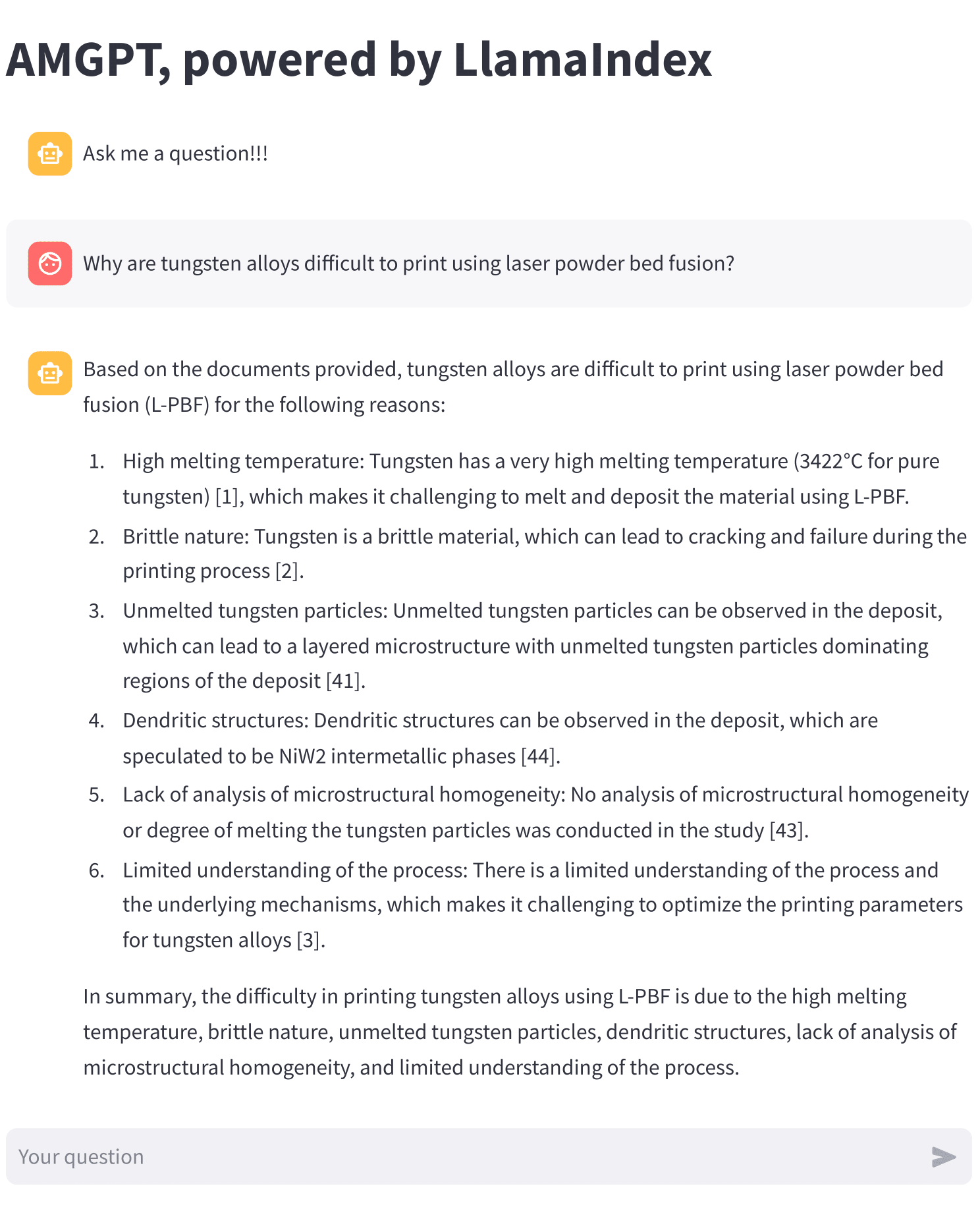}
\caption{Chat Interface for AMGPT built using Streamlit}
\label{fig:chat_stmlit}
\end{figure}

Tools like Streamlit\footnote{Streamlit is available at \url{https://github.com/streamlit/streamlit}} simplify developing interactive web applications, enabling quick creation of user interfaces for machine learning and data science projects without extensive web development knowledge. As an open-source Python library, Streamlit facilitates this process. In this research, Streamlit was used to build a user-friendly chat interface as shown in Figure \ref{fig:chat_stmlit}, allowing users to interact seamlessly with a machine learning model that analyzes input and provides responses. Additionally, Streamlit helps to record chat memory for future prompting, enhancing the continuity and relevance of the interactions.

\newpage

\section{Results} 
\label{sec:results}

\subsection{Influence of sampling temperature}
Sampling temperature \cite{hinton} is a parameter that determines the randomness and hence, the creativity of the output of the LLM during the RAG execution. Lower sampling temperatures ($<0.5$) result in tokens of higher probability being chosen for generated outputs, conversely higher temperatures ($>1$) result in lower probability tokens being chosen and may cause very divergent and/or "hallucinatory" outputs for consecutive executions of the same prompt. A temperature of 0 results in deterministic outputs such that the generated text remains the same for every iteration of the same query. Lower temperatures are ideal for the precise retrieval of facts, while higher temperatures can aid in more creative multidisciplinary explorations of knowledge bases. Using the Llama2-RAG LLM, vague or hallucinatory terms began to appear at a temperature equal to or greater than 1.5. These effects are shown in Table \ref{table:samp_temp} below.

\newpage

\begin{singlespace}
\textbf{\captionof{table}{Effect of Sampling Temperature on Response Quality}\label{table:samp_temp}}
\centering
\small{
\noindent\rule{\textwidth}{0.4pt} 
\vspace{3pt}
\raggedright
\textbf{Query:} Why are high strength aluminum alloys difficult to print in laser powder bed fusion?
\vspace{3pt}

\centering
\begin{tabularx}{\linewidth}{@{}>{\raggedright\arraybackslash}p{0.66\linewidth}X>{\raggedright\arraybackslash}p{0.13\linewidth}X>{\raggedright\arraybackslash}p{0.21\linewidth}X@{}} 

\textbf{Response} & \textbf{Sampling} & \textbf{Comments} \\ 
&\textbf{Temperature}  &\\
\hline
 High strength aluminum alloys are difficult to print in laser powder bed fusion (LPBF) due to several reasons:

1. High strength alloys have a \textbf{higher melting temperature }than other aluminum alloys, which can make them more difficult to melt and solidify using LPBF.
    
2. These alloys often have a more \textbf{complex microstructure}, which can lead to issues during the melting and solidification process, such as porosity and residual stresses.
    
3. The high strength of these alloys can also make them more prone to \textbf{cracking and distortion }during the LPBF process, which can affect their mechanical properties.
    
4. The \textbf{thermal conductivity }of high strength alloys is often lower than that of other aluminum alloys, which can make it more difficult to achieve uniform melting and solidification.
    
5. The \textbf{laser power and beam profile }can also affect the LPBF process, and may not be suitable for high strength alloys, leading to poor surface finish and mechanical properties.

Overall, the challenges in printing high strength aluminum alloys using LPBF are related to their high melting temperature, complex microstructure, and lower thermal conductivity, as well as the need for optimal laser power and beam profile.
& 0.1 & This is a well-crafted response with some strong points \\\\

\hline
\end{tabularx}
}
\end{singlespace}

\newpage
\begin{singlespace}
\addtocounter{table}{-1}  
\captionof{table}{(continued)} 
\centering
\small{
\noindent\rule{\textwidth}{0.4pt} 
\vspace{3pt}
\raggedright
\textbf{Query:} Why are high strength aluminum alloys difficult to print in laser powder bed fusion? 
\vspace{3pt}

\centering
\begin{tabularx}{\linewidth}{@{}>{\raggedright\arraybackslash}p{0.66\linewidth}X>{\raggedright\arraybackslash}p{0.13\linewidth}X>{\raggedright\arraybackslash}p{0.21\linewidth}X@{}} 

\textbf{Response} & \textbf{Sampling} & \textbf{Comments} \\ 
&\textbf{Temperature}  &\\
\hline

High strength aluminum alloys are difficult to print in laser powder bed fusion (LPBF) due to several reasons. Here are some of the key challenges:

1. \textbf{Higher melting point: }High strength aluminum alloys have a higher melting point than other aluminum alloys, which makes it more difficult to melt and solidify them using LPBF. This can result in slower build times and increased thermal management challenges.
2. \textbf{Difficulty in achieving uniform melting:} The high strength aluminum alloys have a more complex microstructure, which can make it challenging to achieve uniform melting and consolidation during the LPBF process. This can lead to variations in the microstructure and mechanical properties of the final product.
3. \textbf{Sensitivity to laser power and beam profile:} High strength aluminum alloys are sensitive to the laser power and beam profile used in LPBF. If the laser power is too high or the beam profile is not optimal, it can result in overheating, melting, or cracking of the alloy, leading to defects and poor mechanical properties.
4. \textbf{Limited build height: }The build height in LPBF is limited by the powder bed height and the laser beam height. This can make it difficult to print high-strength aluminum alloys with complex geometries or large build volumes.
5. \textbf{Difficulty in achieving uniform cooling:} Uniform cooling is essential to achieve consistent mechanical properties in LPBF. However, high strength aluminum alloys can be prone to uneven cooling due to their complex microstructure, which can lead to variations in the mechanical properties of the final product.

Overall, the challenges in printing high strength aluminum alloys using LPBF are related to their higher melting point, complex microstructure, sensitivity to laser power and beam profile, limited build height, and difficulty in achieving uniform cooling. Addressing these challenges can require careful process optimization and the development of new techniques and technologies.
& 0.4 & This response starts meandering slightly, with points like "limited build height" that are irrelevant to the question at hand. This is the effect of a higher sampling temperature.\\\\

\hline
\end{tabularx}
}
\end{singlespace}

\newpage
\begin{singlespace}
\addtocounter{table}{-1}  
\captionof{table}{(continued)} 
\centering
\small{
\noindent\rule{\textwidth}{0.4pt} 
\vspace{3pt}
\raggedright
\textbf{Query:} Why are high strength aluminum alloys difficult to print in laser powder bed fusion? 
\vspace{3pt}

\centering
\begin{tabularx}{\linewidth}{@{}>{\raggedright\arraybackslash}p{0.66\linewidth}X>{\raggedright\arraybackslash}p{0.13\linewidth}X>{\raggedright\arraybackslash}p{0.21\linewidth}X@{}} 

\textbf{Response} & \textbf{Sampling} & \textbf{Comments} \\ 
&\textbf{Temperature}  &\\
\hline

High strength aluminum alloys are difficult to print in laser powder bed fusion (LPBF) due to several challenges. Firstly, these alloys have a high strength-to-weight ratio, which can make them prone to \textbf{cracking and distortion} during the melting process. This is because the high thermal conductivity of the alloy can cause rapid solidification, leading to stresses and distortions that can result in defects and cracking.

Secondly, high strength aluminum alloys often have a \textbf{complex microstructure}, which can make it difficult to achieve uniform melting and consolidation. This can lead to variations in the microstructure and mechanical properties of the final product, which can affect its performance and reliability.

Thirdly, the high temperature required for LPBF can cause \textbf{oxidation and contamination }of the melt pool, which can affect the mechanical properties of the final product. This is particularly true for high strength aluminum alloys, which have a high susceptibility to oxidation and contamination.

Finally, the limited build height and layer thickness in LPBF can make it difficult to print\textbf{ large or complex parts }with high strength aluminum alloys. This can require the use of specialized build strategies or the use of multiple builds to achieve the desired part geometry.

Overall, the challenges associated with printing high strength aluminum alloys in LPBF highlight the need for careful process optimization and material selection to achieve high-quality parts with the desired properties
& 0.7 & Sometimes a higher sampling temperature can have the positive effect of highlighting good points like "oxidation and contamination"\\\\

\hline
\end{tabularx}
}
\end{singlespace}

\newpage
\begin{singlespace}
\addtocounter{table}{-1}  
\captionof{table}{(continued)} 
\centering
\small{

\noindent\rule{\textwidth}{0.4pt} 
\vspace{3pt}
\raggedright
\textbf{Query:} Why are high strength aluminum alloys difficult to print in laser powder bed fusion? 
\vspace{3pt}

\centering
\begin{tabularx}{\linewidth}{@{}>{\raggedright\arraybackslash}p{0.66\linewidth}X>{\raggedright\arraybackslash}p{0.13\linewidth}X>{\raggedright\arraybackslash}p{0.21\linewidth}X@{}} 

\textbf{Response} & \textbf{Sampling} & \textbf{Comments} \\ 
&\textbf{Temperature}  &\\
\hline

The article discusses the challenges of printing high strength aluminum alloys using laser powder bed fusion (LPBF). The main reasons for this difficulty are:

1. \textbf{High melting temperature: }Aluminum alloys have a high melting temperature, which can lead to \textcolor{red}{poor fusion} of the particles during the LPBF process, resulting in porosity and poor strength.
2. \textbf{Low thermal conductivity: }The thermal conductivity of aluminum alloys is low, which can result in slow heat transfer during the LPBF process, leading to inconsistent temperature distribution and poor quality.
3. High sensitivity to process parameters: Aluminum alloys are highly sensitive to process parameters such as laser power, scan speed, and beam overlap, which can affect their microstructure and properties.
4. \textbf{Limited available alloys:} There are limited alloys available for LPBF that offer the required strength and durability for high-performance applications.
5. \textbf{Cost and complexity: }LPBF is a costly and complex process, and the availability of high-strength aluminum alloys may be limited due to cost and manufacturing constraints.

These factors make it challenging to print high strength aluminum alloys using LPBF, but various approaches are being explored to overcome these challenges, such as the use of alternative alloys, optimization of process parameters, and the development of new manufacturing techniques.
& 1.5 &Strange language like "poor fusion" starts appearing at higher temperatures.\\\\

\hline
\end{tabularx}
}
\end{singlespace}

\newpage

\subsection{Influence of system prompt}
The system prompt is a set of instructions provided to the large language model (LLM) before the knowledge retrieval step. These instructions give users a certain level of qualitative control over the descriptive nature of the generated response. The system prompt can be used for content moderation and to modulate the response length and type of language used, ensuring a coherent output \cite{10.1145/3560815}. Our retrieval-augmented generation (RAG) system accurately follows system prompts, demonstrating that the pre-trained model possesses natural language understanding capabilities.

For instance, when the system is pre-prompted to adjust its responses from excluding offensive words and personal anecdotes to delivering explanations suitable for a 10-year-old, it avoids complex terminology typically used in university-level discussions, such as "gradients", "optimization", and "dendritic structures." Instead, the system simplifies its language to ensure that the explanations remain accessible and easy to comprehend for a general audience. These effects are shown in Table \ref{table:sys_prompt} below.

\newpage

\begin{singlespace}
\textbf{\captionof{table}{Effect of System Prompt}\label{table:sys_prompt}}
\centering
\small{
    \noindent\rule{\textwidth}{0.4pt} 
    \vspace{3pt}
    \raggedright
    \textbf{Query:} What is crystallographic texture? Find in the literature, what crystallographic texture can often be observed in L-PBF b-Ti alloys? Suggest ways to manipulate texture in L-PBF materials.
    \centering
    \begin{tabularx}{\linewidth}{@{}>{\raggedright\arraybackslash}p{0.26\linewidth}X>{\raggedright\arraybackslash}p{0.14\linewidth}X>{\raggedright\arraybackslash}p{0.6\linewidth}X@{}} 
        \textbf{System Prompt} & \textbf{Response} & \textbf{Comments} \\ 
        \hline
\small{
    You are an AI assistant that answers questions in a friendly manner, based on the given source documents.

    \textbf{-} Generate human readable output, avoid creating output with gibberish text.

    \textbf{-} Generate only the requested output, don't include any other language before or after the requested output.

    \textbf{-} Never say thank you, that you are happy to help, that you are an AI agent, etc. Just answer directly.

    \textbf{-} Generate professional language.

    \textbf{-} Never generate offensive or foul language.

    \textbf{-} Do not write "The authors" in any answer.

    \textbf{-} Do not use "[]" in any answer.

    \textbf{-} Write every answer like a list of known facts without referring to anybody or any document in the third person.

    \textbf{-} Never use references in square brackets or otherwise in the output, but provide material examples if possible.
}

        &
        Crystallographic texture refers to the orientation of crystals in a material, which can affect the physical and mechanical properties of the material. In L-PBF b-Ti alloys, crystallographic texture can often be observed due to the epitaxial nucleation at the substrate or previously melted and solidified layer, resulting in colonies of cells/dendrites having the same crystallographic orientation. This crystallographic texture can be inherited from layer to layer, developing a texture.

        To manipulate texture in L-PBF materials, several approaches can be taken:
        1. \textbf{Change in laser scanning strategy:} By altering the laser scanning strategy, the solidification texture can be minimized. For example, using a slower laser speed or a different scanning direction can help reduce the texture.
        2. \textbf{Preheating temperature:} The preheating temperature of the build platform can be adjusted to influence the microstructure and texture of the material. Higher preheating temperatures can lead to a more equiaxed microstructure and reduce the crystallographic texture.
        3. \textbf{Heat treatment:} The material can be subjected to a heat treatment to alter the microstructure and texture. This can help reduce the crystallographic texture and achieve a more isotropic material.
        4. \textbf{Material selection:} Selecting materials with a specific crystal structure or microstructure can help manipulate the texture. For example, using materials with a cubic crystal structure can result in a more isotropic material.
        5. \textbf{Genetic algorithms and machine learning:} These computer-aided techniques can be used to optimize process parameters and manipulate the texture of L-PBF materials. By analyzing the relationship between process parameters and microstructure, these algorithms can identify the optimal parameters for a specific material and texture.
        & This information is largely accurate and acceptable as a solid summary.\\
        \hline
    \end{tabularx}
}
\end{singlespace}

\newpage
\begin{singlespace}
\addtocounter{table}{-1}  
\captionof{table}{(continued)} 
\centering
\small{
\noindent\rule{\textwidth}{0.4pt} 
\vspace{3pt}
\raggedright
\textbf{Query:} What is crystallographic texture? Find in the literature, what crystallographic texture canoften be observed in L-PBF b-Ti alloys? Suggest ways to manipulate texture in L-PBF materials.?
\vspace{3pt}
\begin{tabularx}{\linewidth}{@{}>{\raggedright\arraybackslash}p{0.26\linewidth}X>{\raggedright\arraybackslash}p{0.14\linewidth}X>{\raggedright\arraybackslash}p{0.7\linewidth}X@{}} 
\textbf{System Prompt} & \textbf{Response} & \textbf{Comments} \\ 
\hline
You are an expert on

additive manufacturing

that answers questions in

a friendly manner, based

on the given source

documents. Here are

some rules you always

follow:

- Generate human

readable output, avoid

creating output with

gibberish text.

- Keep your answers very

brief

- Do not refer to any

documents, figures in

your answer. just give

me the answer that you

extract from them.

-Never use references in

square brackets or

otherwise in the output,

but provide material

examples if possible
&
Crystallographic texture refers to the orientation of crystals in a material, which can affect the physical and mechanical properties of the material. In L-PBF (laser powder bed fusion) materials, crystallographic texture can arise due to the epitaxial nucleation at the substrate or previously melted and solidified layer, leading to the formation of colonies of cells or dendrites with the same crystallographic orientation. This texture can be influenced by factors such as laser scanning strategy, preheating temperature, and scanning direction.
In L-PBF b-Ti alloys, crystallographic texture can often be observed in the form of a weak HCP (hexagonal close-packed) texture, which is retained even after incomplete relaxation annealing at 650 \degree C for 3h. This texture can be minimized by variations of scanning strategies and change of scanning direction from layer to layer.
Ways to manipulate texture in L-PBF materials include:
1. \textbf{Laser scanning strategy:} Changing the laser scanning strategy can alter the texture in L-PBF materials. For example, using a slower laser speed can result in a coarser grain size and a more uniform texture.
2. \textbf{Preheating temperature: }The preheating temperature can also influence the texture in L-PBF materials. Higher preheating temperatures can lead to a finer grain size and a more uniform texture.
3.\textbf{ Scanning direction:} Changing the scanning direction from layer to layer can also manipulate the texture in L-PBF materials. For example, scanning in a specific direction can result in a more uniform texture.
4.\textbf{ Heat treatment:} Heat treatment can also be used to manipulate the texture in L-PBF materials. For example, a selective heat treatment can be used to reduce the amount of acuity in the texture.
5.\textbf{ Material selection:} Selecting materials with specific properties can also manipulate the texture in L-PBF materials. For example, using materials with a higher melting temperature can result in a more uniform texture. 
&
This is a more comprehensive, expert-style answer which is kept brief. However, the use of the word "acuity" in point 4 is uncommon. \\
\hline
\end{tabularx}
}
\end{singlespace}
\newpage
\begin{singlespace}
\addtocounter{table}{-1}  
\captionof{table}{(continued)} 
\centering
\small{
\noindent\rule{\textwidth}{0.4pt} 
\vspace{3pt}
\raggedright
\textbf{Query:} What is crystallographic texture? Find in the literature, what crystallographic texture can often be observed in L-PBF b-Ti alloys? Suggest ways to manipulate texture in L-PBF materials.?
\vspace{3pt}

\begin{tabularx}{\linewidth}{@{}>{\raggedright\arraybackslash}p{0.26\linewidth}X>{\raggedright\arraybackslash}p{0.14\linewidth}X>{\raggedright\arraybackslash}p{0.7\linewidth}X@{}} 
 \textbf{System Prompt} & \textbf{Response} & \textbf{Comments}\\ 
 \hline
You are a science and technology populariser who seeks to explain concepts in a simple manner. 
&
Crystallographic texture refers to the orientation of crystals in a material, which can affect the physical and mechanical properties of the material. In L-PBF b-Ti alloys, crystallographic texture can often be observed due to the epitaxial nucleation at the substrate or previously melted and solidified layer, leading to the formation of colonies of cells/dendrites having the same crystallographic orientation. These colonies that have favorable orientation and conditions of highest growth rate will dominate the microstructure, and this microstructure is inherited from layer to layer, developing a texture.

To manipulate texture in L-PBF materials, several approaches can be taken:

1. \textbf{Change and control of process parameters:} Preheating temperature, scanning strategy, and laser power can influence the texture formation in L-PBF materials. By altering these parameters, the texture can be altered, and the acuity of the texture can be reduced.
2.\textbf{ Heat treatment: }Heat treatment can be used to alter the microstructure and texture of L-PBF materials. By applying different heat treatment conditions, the crystallographic texture can be controlled, and the material's properties can be tailored.
3. \textbf{Selection of scanning strategy: }The scanning strategy used in L-PBF can affect the texture formation. By selecting appropriate scanning strategies, such as layer-by-layer or track-by-track, the texture can be minimized, and the material's properties can be tailored.
4.\textbf{ Use of nanostructured materials:} Nanostructured materials can exhibit unique texture and microstructure properties due to their small size. By using nanostructured materials in L-PBF, the texture can be manipulated, and the material's properties can be tailored.
5. \textbf{Use of additives:} Additives can be used to modify the microstructure and texture of L-PBF materials. By adding certain additives, the crystallographic texture can be controlled, and the material's properties can be tailored.

&
This response is a very simplified one that seeks to explain concepts to a layman. \\
\hline
\end{tabularx}
}
\end{singlespace}

\newpage

\subsection{Influence of max token length}
The max token length is a parameter that dictates the maximum of number of tokens that the LLM can use while generating a response. A short max token length ($<200$) forces the LLM to produce a response of a few brief sentences at most. A larger max token length ($>256$) achieves greater response lengths which are useful for very esoteric and descriptive answers. Very straightforward and fact-based one-line answers are unaffected by the max token length. Long token lengths ($\geq1024$) caused the system to hallucinate after running out of relevant information to incorporate in its answer. The responses tabulated in this section use a max token length of 768.

\subsection{\text{top\_k} Retrieval Methodology}
In the context of embedding-based retrieval, \text{top\_k} \cite{xiao} is a critical parameter that determines the number of top embeddings considered during the search process. It specifies the number of closest matches (in terms of embedding similarity) that are retrieved from the index. A larger \text{top\_k} value implies a broader search scope, potentially enhancing the accuracy of the results by considering more candidates. However, this increase in scope needs to be balanced against computational efficiency and the limitations imposed by system resources, such as the max token length parameter, which defines the maximum allowable size of the token set for processing. 

The similarity measurement used is cosine similarity \cite{Gunawan_2018}, which calculates the cosine of the angle between the query embedding and each indexed embedding. This is done by taking the inner product of the query vector and an indexed vector, normalized by the product of their magnitudes. This normalization allows the similarity calculation to focus solely on the directionality of the vectors, thus making it scale-invariant and particularly suited for high-dimensional data comparisons.
Cosine similarity scores range from -1 (perfect dissimilarity) to 1 (perfect similarity), with 0 indicating no similarity. The search process retrieves items with the highest positive cosine similarity scores, effectively identifying the most relevant items in the index.

For the Llama2-RAG LLM, a topk value of 3 produced the most balanced response by more examples than a topk of 2 without digressing when topk was 4 or 6. The effects are shown in Table \ref{table:topk} below. 

\begin{singlespace}
\textbf{\captionof{table}{Effect of Similarity top\_k}\label{table:topk}}
\centering
\small{
\noindent\rule{\textwidth}{0.4pt} 
\vspace{3pt}
\raggedright
\textbf{Query:} Give examples of nickel-based superalloys and common application areas for them?
\vspace{3pt}
\begin{tabularx}{\linewidth}{@{}>{\raggedright\arraybackslash}p{0.6\linewidth}>{\raggedright\arraybackslash}p{0.22\linewidth}>{\raggedright\arraybackslash}p{0.18\linewidth}@{}}
\textbf{Response} & \textbf{Similarity topk} & \textbf{Comments} \\
\hline
Based on the given context information, nickel-based superalloys are a class of alloys that have been widely used in various applications due to their high-temperature resistance and strength. Some examples of nickel-based superalloys include:
\textbf{IN718:} This is a widely used nickel-based superalloy that is known for its high strength, good corrosion resistance, and high-temperature resistance. It is commonly used in gas turbines, nuclear reactors, and chemical processing equipment.
\textbf{IN625:} This alloy is known for its high strength, good corrosion resistance, and high-temperature resistance. It is commonly used in aerospace applications, such as turbine blades and engine components, due to its ability to withstand high temperatures and corrosive environments.
& 2 & This is a highly focused response with a few examples. \\\\
Based on the provided context information, here are some examples of nickel-based superalloys and their common application areas:
\textbf{IN718:} Used in gas turbines, nuclear power plants, and chemical containers.
\textbf{IN625:} Used in land-based gas turbines, nuclear power plants, and chemical containers.
\textbf{IN630:} Used in chemical containers and nuclear power plants.
\textbf{IN738:} Used in gas turbines and chemical containers.
\textbf{IN833:} Used in chemical containers and nuclear power plants.
These nickel-based superalloys are commonly used in various applications due to their excellent mechanical properties, such as high strength, good ductility, and resistance to corrosion and oxidation. They are particularly useful in high-temperature environments and are often used in critical components where reliability and durability are paramount.
& 3 & This is a response that gives more useful examples because it accesses more embeddings.\\
\hline
\end{tabularx}
}
\end{singlespace}
\newpage
\begin{singlespace}
\addtocounter{table}{-1}  
\captionof{table}{(continued)} 

\centering
\small{
\noindent\rule{\textwidth}{0.4pt} 

\vspace{3pt}
\raggedright
\textbf{Query:} Give examples of nickel-based superalloys and common application areas for them?
\vspace{3pt}

\begin{tabularx}{\linewidth}{@{}>{\raggedright\arraybackslash}p{0.66\linewidth}X>{\raggedright\arraybackslash}p{0.14\linewidth}X>{\raggedright\arraybackslash}p{0.2\linewidth}X@{}} 

\textbf{Response} & \textbf{Similarity topk} & \textbf{Comments} \\
\hline
\small{
Based on the provided context, here are some examples of nickel-based superalloys and their common application areas:

\textbf{IN718:} This nickel-based superalloy is commonly used in gas turbines, nuclear power plants, and chemical containers due to its excellent high-temperature strength, corrosion resistance, and good weldability.
\textbf{IN625: }This nickel-based superalloy is used in aerospace applications, such as jet engine components, due to its excellent high-temperature strength, corrosion resistance, and resistance to oxidation.
\textbf{Hastelloy X: }This nickel-molybdenum alloy is used in high-temperature applications, such as gas turbines, chemical processing equipment, and heat exchangers, due to its excellent resistance to corrosion and oxidation.
\textbf{Inconel 625: }This nickel-chromium-molybdenum alloy is used in a variety of applications, including gas turbines, heat exchangers, and chemical processing equipment, due to its excellent high-temperature strength, corrosion resistance, and resistance to oxidation.
These are just a few examples of nickel-based superalloys and their common application areas. The use of these alloys in more efficient turbines, such as those mentioned in the additional context, is becoming increasingly popular due to their high-temperature strength and resistance to corrosion. For example, IN718 and IN625 are commonly used in \textbf{gas turbines }due to their excellent \textbf{high-temperature strength }and resistance to thermal degradation. Hastelloy X and Inconel 625 are also used in high-temperature applications due to their excellent \textbf{resistance to corrosion and oxidation}.

In addition, the use of nickel-based superalloys in additive manufacturing is becoming more prevalent, as they offer excellent mechanical properties and resistance to corrosion and oxidation. For example, selective laser melting (SLM) of IN718 and IN625 has been shown to produce parts with excellent mechanical properties and surface finish, making them suitable for use in\textbf{ aerospace }and other high-performance applications.

Overall, nickel-based superalloys offer a range of benefits for high-temperature applications, including excellent high-temperature strength, corrosion resistance, and resistance to thermal degradation. Their use in more efficient turbines, additive manufacturing, and other applications is becoming increasingly widespread due to their versatility and performance.}
& \small{4} & \small{This is a much bigger response that digresses slightly as it retrieves embeddings of slightly lower probability.}\\
\hline
\end{tabularx}
}
\end{singlespace}

\newpage
\begin{singlespace}
\addtocounter{table}{-1}  
\captionof{table}{(continued)} 
\centering
\small{
\noindent\rule{\textwidth}{0.4pt} 

\vspace{3pt}
\raggedright
\textbf{Query:} Give examples of nickel-based superalloys and common application areas for them?
\vspace{3pt}

\begin{tabularx}{\linewidth}{@{}>{\raggedright\arraybackslash}p{0.66\linewidth}X>{\raggedright\arraybackslash}p{0.14\linewidth}X>{\raggedright\arraybackslash}p{0.2\linewidth}X@{}} 

\textbf{Response} & \textbf{Similarity topk} & \textbf{Comments} \\
\hline
\small{
Based on the provided context, the original answer can be refined to better answer the query. Here is the refined answer:

The provided context mentions the use of nickel-based superalloys in \textbf{gas turbines, chemical processing equipment, and heat exchangers}, and highlights the importance of understanding the fatigue behavior of these materials. The context also mentions the use of compression tests and creep tests to evaluate the fatigue properties of nickel-based superalloys.

In response to the query, some examples of nickel-based superalloys commonly used in these applications include \textbf{IN718, IN625, Hastelloy X, Monel 400, and Inconel 625.} These alloys are known for their high-temperature strength, corrosion resistance, and good weldability, making them suitable for use in harsh environments.

The context also mentions the use of \textbf{etchants} to prepare nickel-based superalloys for analysis. Common etchants used in literature include oxalic acid, phosphoric acid, and hydroxypropionic acid.

In summary, the refined answer is:

Examples of nickel-based superalloys commonly used in gas turbines, chemical processing equipment, and heat exchangers include IN718, IN625, Hastelloy X, Monel 400, and Inconel 625. These alloys are known for their \textbf{high-temperature strength, corrosion resistance, and good weldability}, making them suitable for use in harsh environments. The use of etchants is important for preparing these alloys for analysis.
}
& \small{6} & \small{This answer digresses a lot and starts lecturing about "etchants". This is irrelevant to the question, but it is retrieved based on lower probability. The quality is manifestly worse than the previous ones.}\\

\hline
\end{tabularx}
}
\end{singlespace}

\newpage

\subsection{Evaluation}
Evaluating responses generated by LLMs is a challenging task that may involve fact-checking known literature \cite{wang2023survey}. The retrieval augmented generation response evaluation is performed by a system, where human researchers evaluate responses and rank them according to relevance and accuracy. Subjective remarks are also provided to explain the merits and shortcomings of the generation from the LLM. To ensure impartiality, this evaluation is conducted in a blind manner, meaning the researchers are not informed about whether the response was generated by a standard LLM or a retrieval-augmented LLM (RAG-LLM).

Across queries in Table \ref{table:retrieval_examples}, the Llama2-RAG LLM produced factual responses without hallucinating for 80\% of the prompts compared to 86.7\% by GPT-4. However, GPT-4 produced a vague response to one query as assessed by a human expert whereas the RAG system maintained a high level of specificity. The word limit for GPT-4's responses was capped at 300 words to enable fair comparison with the RAG system. Am exhaustive set of example responses is provided in the \hyperref[sec:appendix]{Appendix}.

\begin{singlespace}

\textbf{\captionof{table}{Summary of Important parameters that determine model behaviour}}
\centering
{\scriptsize

\begin{tabular}{p{0.3\linewidth} p{0.3\linewidth} p{0.4\linewidth}}
\hline
\hline

\textbf{Parameter} & \textbf{Function} & \textbf{Effects} \\ 
 \hline
 Sampling temperature & Determines tokens for generated response based on probability & A high sampling temperature (e.g. 1.5) leads to creative but often nonsensical responses, while a low sampling temperature (e.g. 0.5) results in safe but uninteresting responses. \\\\
 System prompt & Pre-condition for the model for moderating/controlling the quality of output & A poorly crafted system prompt, "Tell me about...", results in vague responses, while a well-designed system prompt, "Describe the origins of...", leads to informative and engaging responses. \\\\
 Max token length & Determines the maximum number of tokens that can be used while generating a response & A short max token length (e.g. 50) causes brief, incomplete responses, while a longer max token length (e.g. 200) allows for more detailed and coherent responses. \\\\
 Similarity topk & Selecting the top k embeddings most similar to a query for use in retrieval-augmented generation & A low similarity topk value (e.g. 5) results in generated responses that are overly similar, while a higher similarity topk value (e.g. 20) leads to responses that are more varied and interesting. \\\\
 Prompt & Query given to the LLM to extract responses from the knowledge corpus & A vague prompt, "What is AI?", yields responses that are too broad, while a specific prompt, "Explain the differences between transformers and recurrent neural networks", leads to detailed and informative responses. \\\\
 \hline
\end{tabular}

\label{tab:query-generation}
}
\end{singlespace}

\newpage

\section{Conclusion and Future Work}
\label{sec:conclusion}

We characterized RAG methods by varying inference parameters to produce a reliable metal additive manufacturing expert LLM that can be queried through a user interface. Due to the nature of the corpus-referencing task in constraining a response to be consistent with an external factual base, minimizing the topk and temperature parameters yielded the most relevant results. Despite having less than 0.5\% of the parameters that GPT-4 has, our RAG system is able to maintain high fidelity and accuracy of answers. RAG efficiently enhances the question-answering capabilities of LLMs.

A paper by Melz \cite{melz2023enhancing} proposes ARM-RAG, which deploys an information retrieval model with a vector database called FAISS, developed by Facebook, and employs a maximum inner product search (MIPS) to extract information. It shows the process of sequentially querying an LLM for auxiliary answers to expand relevant context. This approach could further enhance the capabilities of our system by improving the retrieval process and expanding the relevant context for more accurate responses.

Due to the timeline, hardware limitations, and the need for external subscriptions to LLM services, fine-tuning is not included in the scope of AMGPT but would benefit future exploration. Incorporating techniques such as ARM-RAG and fine-tuning could potentially lead to even greater accuracy and reliability in our LLM-RAG system. 

Future exploration includes advancing the memory capabilities for the chat memory feature. By enhancing how the system records and recalls past interactions, we aim to improve the continuity and context-awareness of responses, leading to a more robust and user-friendly experience. This development will be crucial in maintaining the relevance and accuracy of information over prolonged interactions.

Additionally, future research will focus on improving the quality and detail of image descriptions in journal papers. By leveraging advanced image recognition and natural language processing techniques, we can generate more precise and informative descriptions of complex images and diagrams. This will not only enhance the clarity and accessibility of visual data in academic publications but also support more effective communication of research findings.

\section*{\textbf{Acknowledgments}}
\label{sec:acknowledgements}

We gratefully acknowledge the insightful discussions and valuable evaluations provided by Peter Pak and Abraham George from the Mechanical and AI Lab at Carnegie Mellon University, and Barnali Mondal from the Malen Laboratory at Carnegie Mellon University.



 \bibliographystyle{elsarticle-num} 
 \bibliography{cas-refs}





\newpage
\section{Appendix}
\label{sec:appendix}
\begin{singlespace}
\textbf{\captionof{table}{Examples to illustrate retrieval}\label{table:retrieval_examples}} 
\noindent\rule{\textwidth}{0.4pt} 

\vspace{3pt}
\raggedright
\textbf{Query:} What is electrical discharge machining?\\
\vspace{3pt}
\begin{tabularx}{\linewidth}{@{}>{\raggedright\arraybackslash}p{0.4\linewidth}X>{\raggedright\arraybackslash}p{0.15\linewidth}X>{\raggedright\arraybackslash}p{0.4\linewidth}X@{}}

\hline
\textbf{LLAMA2-7B RAG} & \textbf{GPT-4} & \textbf{Comments} \\ 
\hline
\small{

Electrical discharge machining (EDM) is a manufacturing process that uses \textbf{electrical discharges to remove material from a workpiece.} It is a non-traditional machining process that can be used to produce parts with complex geometries and high accuracy. In EDM, a electrical discharge is created between two electrodes immersed in a dielectric fluid, and the discharge vaporizes the workpiece material, removing it from the electrodes. The process can be used to machine a wide range of materials, including metals, ceramics, and composites. EDM is often used in situations where traditional machining processes are difficult or impossible to use, such as machining small parts with complex shapes or machining materials that are difficult to cut.} & 
\small{
Electrical Discharge Machining (EDM) is a manufacturing process that utilizes \textbf{electrical discharges or sparks to remove material from a workpiece.} It is particularly useful for machining complex shapes in hard materials that are difficult to cut with traditional mechanical tools. The process works by creating an electrical spark between an electrode and the workpiece, both submerged in a dielectric fluid. The spark produces intense heat that melts and vaporizes the material on the surface of the workpiece in a very localized area.

   EDM is divided into two main types:
   - \textbf{Wire EDM}: Uses a continuously moving thin wire as an electrode, typically used to cut plates as thick as 300mm and to make punches, tools, and dies from hard metals that are difficult to machine with other methods.

   - \textbf{Die-sinking EDM}: Uses a machined electrode that forms a mirror image of the desired shape in the workpiece, often used for complex geometries like molds, dies, and cavities.
} & 
\small{ 
While both responses appear to be technically correct, the GPT-4 Response is more detailed than the RAG Response.}\\
\hline
\end{tabularx}
\end{singlespace}

\newpage
\begin{singlespace}
\addtocounter{table}{-1}  
\captionof{table}{(continued)} 

\noindent\rule{\textwidth}{0.4pt} 
\vspace{3pt}
\raggedright
\textbf{Query:} Which NDT methods work well for laser powder bed fusion?\\
\vspace{3pt}
\begin{tabularx}{\linewidth}{@{}>{\raggedright\arraybackslash}p{0.4\linewidth}X>{\raggedright\arraybackslash}p{0.15\linewidth}X>{\raggedright\arraybackslash}p{0.4\linewidth}X@{}}

\hline
\textbf{LLAMA2-7B RAG} & \textbf{GPT-4} & \textbf{Comments} \\ 
\hline
\small{

The following NDT methods are likely to work well for laser powder bed fusion:

1. \textbf{Radiography: }X-ray microCT and radiography are suitable for detecting porosity and other defects in laser powder bed fusion parts.
2. \textbf{Ultrasonic:} Ultrasonic techniques can be used to detect surface cracks and other defects in laser powder bed fusion parts.
3. \textbf{Magnetic particle inspection:} Magnetic particle inspection can be used to detect surface cracks and other defects in laser powder bed fusion parts.
4. \textbf{Eddy-current:}Eddy-current testing can be used to detect surface cracks and other defects in laser powder bed fusion parts.
5. \textbf{Liquid penetrant:}Liquid penetrant testing can be used to detect surface cracks and other defects in laser powder bed fusion parts.

It is important to note that the suitability of these NDT methods may depend on the specific application, the size and complexity of the part, and the desired level of accuracy. } &
\small{
   Non-Destructive Testing (NDT) methods that are effective for evaluating parts made by Laser Powder Bed Fusion include:
   - \textbf{X-ray Computed Tomography (CT): }Provides detailed images of internal structures, detecting pores, cracks, and other defects. It is especially valuable for its ability to assess the internal integrity and the exact dimensions within complex geometries.
   - \textbf{Ultrasonic Testing:}Useful for detecting delaminations and inconsistencies in density. Ultrasonic waves are sent through the material and reflections are analyzed to identify flaws.
  \textcolor{red}{ Thermal Imaging: Involves monitoring the thermal emissions from the part during or after production. Variations in thermal conductivity can indicate anomalies such as cracks or poor fusion between layers.} } &
\small{
The GPT-4 response refers to thermal imaging as a non-destructive testing method, but this is an in-situ monitoring method. Also, variations in thermal conductivity is not directly provided by in-situ thermal imaging. The RAG response correctly identifies multiple methods of non-destructive testing.
}\\\\
\hline
\end{tabularx}
\end{singlespace}

\newpage
\begin{singlespace}
\addtocounter{table}{-1}  
\captionof{table}{(continued)} 
\noindent\rule{\textwidth}{0.4pt} 
\vspace{3pt}
\raggedright
\textbf{Query:} Which materials are more prone to form keyhole defects in laser powder bed fusion?\\
\vspace{3pt}
\begin{tabularx}{\linewidth}{@{}>{\raggedright\arraybackslash}p{0.4\linewidth}X>{\raggedright\arraybackslash}p{0.2\linewidth}X>{\raggedright\arraybackslash}p{0.4\linewidth}X@{}}

\hline
\textbf{LLAMA2-7B RAG} & \textbf{GPT-4} & \textbf{Comments} \\ 

\hline
\small{
Materials that are more prone to form keyhole defects in laser powder bed fusion are those with high thermal conductivity and high melting temperatures. This is because keyhole defects occur when the laser energy is concentrated in a small area, causing the material to vaporize and leave behind a hole. Materials with high thermal conductivity and high melting temperatures tend to conduct heat more efficiently, which can lead to more \textcolor{red}{intense heat concentration} and a higher likelihood of keyhole defects.

Some examples of materials that may be more prone to keyhole defects in laser powder bed fusion include:

- Metals with high thermal conductivity, such as aluminum and titanium
- Materials with high melting temperatures, such as stainless steel and nickel-based alloys
- Materials with a \textcolor{red}{high density and low porosity, as these can reduce the likelihood of pores forming} during the melting process.} & 
\small{
Keyhole defects in L-PBF are typically more prevalent in materials with high reflectivity and high thermal conductivity. These characteristics can cause unstable melt pool dynamics due to the material’s tendency to reflect laser energy and rapidly conduct heat away from the melt zone. Materials commonly associated with keyhole defects include:
    - \textbf{Titanium and its alloys:} Known for their high thermal conductivity and sensitivity to temperature variations, making them susceptible to keyhole defects.
    - \textbf{Aluminum alloys:}Their high thermal conductivity and low viscosity at high temperatures make them prone to such defects, especially when improper parameters are used.
    - Materials like \textbf{copper alloys} also pose challenges due to their high reflectivity, which can lead to insufficient absorption of laser energy, promoting keyhole defect formation. } & 
\small{ 
The GPT-4 response (Response 2) is more accurate than the RAG response. (Response 1). Response 1 correctly identifies the high-level reasons why keyhole defects are more likely in certain materials, but contains a few inaccurate details. For instance, a high conductivity would lead to a less intense heat concentration due to heat dissipation. Additionally, the porosity of the powder itself is not a material property, and is separate from the mechanism of keyhole porosity formation.}\\
\hline
\end{tabularx}
\end{singlespace}

\newpage
\begin{singlespace}
\addtocounter{table}{-1}  
\captionof{table}{(continued)} 
\noindent\rule{\textwidth}{0.4pt} 
\vspace{3pt}
\raggedright
\textbf{Query:} Why are tungsten alloys difficult to print using laser powder bed fusion?\\
\vspace{3pt}
\begin{tabularx}{\linewidth}{@{}>{\raggedright\arraybackslash}p{0.4\linewidth}X>{\raggedright\arraybackslash}p{0.15\linewidth}X>{\raggedright\arraybackslash}p{0.4\linewidth}X@{}}

\hline
\textbf{LLAMA2-7B RAG} & \textbf{GPT-4} & \textbf{Comments} \\ 

\hline
\small{
Tungsten alloys are difficult to print using laser powder bed fusion due to several reasons:

\textbf{High melting temperature: }Tungsten has a very high melting temperature (3422°C for pure tungsten), which makes it challenging to melt and form a coherent material using laser powder bed fusion.
\textbf{Brittle nature: }Tungsten is a brittle material, which means it can break or shatter easily. This can lead to defects in the printed material, such as cracks and porosity.
\textbf{Limited cooling rate:} The cooling rate during laser powder bed fusion is limited, which can result in residual stresses and cracking in the printed material.
\textbf{Difficulty in achieving uniform melting: }It can be challenging to achieve uniform melting of tungsten particles during laser powder bed fusion, which can lead to non-uniform microstructural properties in the printed material.
\textbf{Particle agglomeration:}Tungsten particles can agglomerate during the printing process, which can lead to non-uniform microstructure in the printed material.
Overall, tungsten alloys are more challenging to print using laser powder bed fusion due to their high melting temperature, brittle nature, and limited cooling rate.} &
\small{
Tungsten alloys present several challenges when printed using laser powder bed fusion (LPBF) due to their intrinsic material properties and the specific requirements of the LPBF process. Firstly, tungsten has a very \textbf{high melting point (around 3422°C)}, which is significantly higher than that of commonly used metals like steel or aluminum. This requires extremely high laser power to achieve proper melting, which can lead to equipment limitations and increased energy consumption.

Moreover, tungsten's \textbf{high thermal conductivity} and thermal expansion coefficient contribute to rapid cooling and solidification rates following laser exposure. These characteristics often result in significant thermal gradients within the build, leading to \textbf{residual stresses}, \textbf{distortion}, and even \textbf{cracking} during the printing process. The material's stiffness and brittleness further exacerbate these issues, as they can increase susceptibility to cracking under thermal stress.

Finally, the \textbf{high density of tungsten}poses challenges in achieving full density in prints without defects like porosity or inclusions, complicating post-processing and potentially affecting the mechanical properties of the final product.} & 
\small{
Both responses are reasonably accurate with the RAG response giving more detail.}\\
\hline
\end{tabularx}
\end{singlespace}
\newpage

\begin{singlespace}
\addtocounter{table}{-1}  
\captionof{table}{(continued)} 
\noindent\rule{\textwidth}{0.4pt} 
\vspace{3pt}
\raggedright
\textbf{Query:} How is phase composition of alloys analysed?\\
\begin{tabularx}{\linewidth}{@{}>{\raggedright\arraybackslash}p{0.4\linewidth}X>{\raggedright\arraybackslash}p{0.15\linewidth}X>{\raggedright\arraybackslash}p{0.4\linewidth}X@{}}

\hline
\textbf{LLAMA2-7B RAG} & \textbf{GPT-4} & \textbf{Comments} \\ 

\hline
\small{
Various techniques can analyze the phase composition of alloys:

1. \textbf{X-ray diffraction (XRD):} Measures X-ray reflections to find crystal structure and phase composition.
2. \textbf{Scanning electron microscopy (SEM) with EDS:} Shows the alloy's morphology and elemental makeup.
3.\textbf{ Transmission electron microscopy (TEM):} Analyzes the crystal structure and phase size at the nanoscale.
4. \textbf{Atomic force microscopy (AFM):} Shows the nanoscale topography and chemical composition of phases.
5. \textbf{Thermogravimetry (TGA): }Identifies thermal behavior and phase transitions.
6. \textbf{Differential scanning calorimetry (DSC):} Measures heat capacity and phase changes.
7.\textbf{ Micro-Raman spectroscopy: }Analyzes the chemical composition and vibrational modes.
8. \textbf{Infrared (IR) spectroscopy:} Finds chemical composition and bonding.

These methods can be used together to understand the alloy's phase composition.} & 
\small{
Alloy phase composition is analyzed using several methods to find the material's phases and distribution:

1.\textbf{ X-ray Diffraction (XRD): }Identifies phases by unique diffraction patterns.
2. \textbf{Scanning Electron Microscopy (SEM):} Provides surface images, revealing morphology.
3. \textbf{Transmission Electron Microscopy (TEM): }Reveals atomic-level structure, useful for phase boundaries.
4. \textbf{Differential Scanning Calorimetry (DSC): }Measures thermal properties, detects transformations.
5. \textbf{Neutron Diffraction:} Penetrates deeply for comprehensive bulk phase analysis.

A combination of these methods is often used to fully understand the alloy's phase composition.} & 
\small{
The RAG Response is more detailed than the GPT-4 response, but both responses appear to be factually correct.}\\
\hline
\end{tabularx}
\end{singlespace}
\newpage
\begin{singlespace}
\addtocounter{table}{-1}  
\captionof{table}{(continued)} 
\noindent\rule{\textwidth}{0.4pt} 
\vspace{3pt}
\raggedright
\textbf{Query:} Why are titanium alloys popular in additive manufacturing?\\
\begin{tabularx}{\linewidth}{@{}>{\raggedright\arraybackslash}p{0.4\linewidth}X>{\raggedright\arraybackslash}p{0.15\linewidth}X>{\raggedright\arraybackslash}p{0.4\linewidth}X@{}}

\hline
\textbf{LLAMA2-7B RAG} & \textbf{GPT-4} & \textbf{Comments} \\ 

\hline

\small{
Titanium alloys are popular in additive manufacturing due to their\textbf{ high strength, light weight, and excellent bio-mechanical properties.} They are also \textbf{corrosion-resistant }and have a \textbf{high yield strength} with high operating temperatures, making them suitable for various industrial applications such as biomedical implants and aerospace components. Additionally, titanium alloys have a\textbf{ highest tensile strength and hardness}, which makes them ideal for high-performance applications.} & 
\small{
Titanium alloys are highly valued in additive manufacturing due to their \textbf{exceptional strength-to-weight ratio, excellent corrosion resistance, and biocompatibility}. These properties make them ideal for aerospace, automotive, and medical applications. Additionally, titanium's compatibility with additive manufacturing technologies like Direct Metal Laser Sintering (DMLS) allows for the production of complex geometries that are difficult to achieve with traditional manufacturing methods, promoting design innovation and material efficiency.} & 
\small{
Both responses appear to be factually correct, but GPT-4 produces a more vague response.}\\

\hline
\end{tabularx}
\end{singlespace}
\newpage

\begin{singlespace}
\addtocounter{table}{-1}  
\captionof{table}{(continued)} 
\noindent\rule{\textwidth}{0.4pt} 
\vspace{3pt}
\raggedright
\textbf{Query:} Why are high strength aluminum alloys difficult to print in laser powder bed fusion?\\
\begin{tabularx}{\linewidth}{@{}>{\raggedright\arraybackslash}p{0.4\linewidth}X>{\raggedright\arraybackslash}p{0.15\linewidth}X>{\raggedright\arraybackslash}p{0.4\linewidth}X@{}}

\hline
\textbf{LLAMA2-7B RAG} & \textbf{GPT-4} & \textbf{Comments} \\ 

\hline

\small{
High strength aluminum alloys are difficult to print in laser powder bed fusion (LPBF) due to several reasons:

1. High strength alloys have a\textbf{ higher melting temperature} than other aluminum alloys, which can make them more difficult to melt and solidify using LPBF.
2. These alloys often have a\textbf{ more complex microstructure}, which can lead to issues during the melting and solidification process, such as porosity and residual stresses.
3. The high strength of these alloys can also make them more prone to \textbf{cracking and distortion d}uring the LPBF process, which can affect their mechanical properties.
4. The \textbf{thermal conductivity }of high strength alloys is often \textbf{lower }than that of other aluminum alloys, which can make it more difficult to achieve uniform melting and solidification.
5. The laser power and beam profile can also affect the LPBF process, and may not be suitable for high strength alloys, leading to\textbf{ poor surface finish }and mechanical properties.

Overall, the challenges in printing high strength aluminum alloys using LPBF are related to their high melting temperature, complex microstructure, and lower thermal conductivity, as well as the need for optimal laser power and beam profile. }& 
\small{
High strength aluminum alloys are challenging to print in laser powder bed fusion (LPBF) primarily due to their\textbf{ high thermal conductivity and low laser absorptivity.} These materials, such as 7075 and 2024 series, quickly dissipate heat during the printing process. This rapid heat loss can lead to uneven cooling and solidification rates across the printed object, increasing the likelihood of defects such as residual stresses, distortions, and even cracking.

Additionally, these alloys are prone to\textbf{ oxidation}, which can affect the quality of the print. During LPBF, the high-energy laser interacts with the powder material, and in the presence of even minute amounts of oxygen, aluminum can form a thin oxide layer. This layer can inhibit proper fusion between the layers, leading to weak bonds and porosity within the printed object.

Furthermore, the composition of high strength aluminum alloys often includes elements that have a higher tendency to react or form \textbf{undesirable phases under rapid cooling conditions }typical of LPBF, complicating the printing process further.} & 
\small{
Both responses are factual. The RAG Response fails to mention reactivity and the GPT-4 response misses out on mentioning the differences in thermal conductivity. Both seem to have different thrusts.}\\
\hline
\end{tabularx}
\end{singlespace}

\newpage
\begin{singlespace}
\addtocounter{table}{-1}  
\captionof{table}{(continued)} 
\noindent\rule{\textwidth}{0.4pt} 
\vspace{3pt}
\raggedright
\textbf{Query:} What do we use the Rosenthal equation for?\\
\vspace{3pt} 
\begin{tabularx}{\linewidth}{@{}>{\raggedright\arraybackslash}p{0.3\linewidth}X>{\raggedright\arraybackslash}p{0.2\linewidth}X>{\raggedright\arraybackslash}p{0.4\linewidth}X@{}}

\hline
\textbf{LLAMA2-7B RAG} & \textbf{GPT-4} & \textbf{Comments} \\ 

\hline

\small{
Based on the given context information, we use the Rosenthal equation to model a moving heat source with a Gaussian distribution. \textcolor{red}{Specifically, the Eagar-Tsai model, which is derived from the Rosenthal equation}, simulates a moving heat source with a Gaussian distribution. The Rosenthal equation is used to solve the temperature distribution at a specified location relative to a point heat source.} & 
\small{
The Rosenthal equation is a fundamental mathematical model used in heat transfer analysis, particularly in welding and laser processing applications like LPBF. The equation helps predict the temperature distribution around a moving heat source, which is critical for understanding:

1. \textbf{Melt Pool Dynamics: }Estimating the size and shape of the melt pool, which influences the microstructure and mechanical properties of the part.

2. \textbf{Thermal Stresses: }Predicting thermal gradients that lead to residual stresses and potential distortions in the manufactured part.

3. \textbf{Process Optimization: }Aiding in the selection and optimization of process parameters such as laser power, speed, and path planning to ensure consistent quality and efficiency.} &
\small{
Comments: Both responses correctly identify the core use cases of the Rosenthal equation. However, the RAG response makes an error in describing the origin of the Eagar-Tsai equation — it is not derived from the Rosenthal equation, it is derived by solving the heat conduction equation with a different heat source formulation.}\\
\hline
\end{tabularx}
\end{singlespace}

\newpage
\begin{singlespace}
\addtocounter{table}{-1}  
\captionof{table}{(continued)} 
\noindent\rule{\textwidth}{0.4pt} 
\vspace{3pt}
\raggedright
\textbf{Query:} What is ductility-dip-cracking in laser powder bed fusion?\\
\vspace{3pt}
\begin{tabularx}{\linewidth}{@{}>{\raggedright\arraybackslash}p{0.4\linewidth}X>{\raggedright\arraybackslash}p{0.15\linewidth}X>{\raggedright\arraybackslash}p{0.4\linewidth}X@{}}

\hline
\textbf{LLAMA2-7B RAG} & \textbf{GPT-4} & \textbf{Comments} \\ 

\hline

\small{
Ductility-dip-cracking is a type of cracking that occurs in laser powder bed fusion (LPBF) during the scanning process. It is a complex phenomenon that involves the interaction of multiple factors, including the material properties, the thermal residual stresses, and the scanning conditions.

In LPBF, the laser beam is scanned across the powder bed, causing the material to melt and fuse together. However, the scanning process can also create \textbf{residual stresses} in the material, which can lead to cracking. Ductility-dip-cracking occurs when the material is in a ductile state and is subjected to a sudden increase in stress, causing it to crack.

The cracking can occur in the form of\textbf{ small cracks or microcracks}, which can propagate along the scan track or in other directions. The size and extent of the cracks can vary depending on the specific conditions of the LPBF process.

To \textbf{mitigate the effects }of ductility-dip-cracking, researchers have proposed various strategies, including\textbf{ preheating the base plate, using a high P/v ratio, and controlling the atmosphere in the process chamber.} These strategies can help to reduce the residual stresses and improve the overall quality of the LPBF process.} &
\small{
Ductility-dip-cracking (DDC) is a type of material failure specific to high-temperature alloys processed by methods like LPBF. DDC occurs due to the \textbf{localized strain }that accumulates during the solidification and cooling phases of the welding or printing process. This strain can lead to\textbf{ microcracks} in regions where the material's ductility is temporarily reduced due to the presence of harmful phases or precipitates at specific temperatures. Factors such as\textbf{ alloy composition, cooling rates, and thermal gradients }significantly influence the susceptibility to DDC. Managing these factors through process optimization and post-processing treatments is crucial to mitigating the risk of DDC in LPBF processes.} &
\small{
Both responses are factually correct and provide reasonable explanations to manage ductility-dip-cracking.}\\
\hline
\end{tabularx}
\end{singlespace}
\newpage
\begin{singlespace}
\addtocounter{table}{-1}  
\captionof{table}{(continued)} 
\noindent\rule{\textwidth}{0.4pt} 
\vspace{3pt}
\raggedright
\textbf{Query:} Give examples of nickel-based superalloys and common application areas for them?\\
\begin{tabularx}{\linewidth}{@{}>{\raggedright\arraybackslash}p{0.45\linewidth}X>{\raggedright\arraybackslash}p{0.15\linewidth}X>{\raggedright\arraybackslash}p{0.4\linewidth}X@{}}

\hline
\textbf{LLAMA2-7B RAG} & \textbf{GPT-4} & \textbf{Comments} \\ 

\hline
\small{
 \textbf{IN718: }This is a widely used nickel-based superalloy that is known for its high strength, good corrosion resistance, and high-temperature resistance. It is commonly used in gas turbines, nuclear reactors, and chemical processing equipment.
\textbf{ IN625:} This alloy is known for its high strength, good corrosion resistance, and high-temperature resistance. It is commonly used in aerospace applications, such as turbine blades and engine components, due to its ability to withstand high temperatures and corrosive environments.

Common application areas for nickel-based superalloys include:

1. \textbf{Gas turbines: }Nickel-based superalloys are widely used in gas turbines due to their high-temperature resistance and strength. They are used in components such as turbine blades, turbine discs, and seals.
2. \textbf{Aerospace: }Nickel-based superalloys are used in aerospace applications such as engine components, turbine blades, and heat shields due to their high strength, good corrosion resistance, and high-temperature resistance.
3. \textbf{Nuclear reactors:} Nickel-based superalloys are used in nuclear reactors due to their ability to withstand high temperatures and corrosive environments. They are used in components such as fuel cladding and control rods.
4. \textbf{Chemical processing equipment: }Nickel-based superalloys are used in chemical processing equipment such as reactors, heat exchangers, and storage tanks due to their high-temperature resistance and corrosion resistance.} & 
\small{
Nickel-based superalloys, such as \textbf{Inconel 718, Inconel 625, Hastelloy, and Rene 41}, are renowned for their excellent mechanical strength and resistance to thermal creep deformation, good surface stability, and resistance to corrosion or oxidation. Common applications include:

- \textbf{Aerospace: }Used in turbine blades, exhaust systems, and engine components due to their ability to withstand extreme environments and high temperatures.
- \textbf{Energy Sector:} Utilized in gas turbine components and nuclear reactor cores.
- \textbf{Oil and Gas Industry:} Employed in piping systems and reactor vessels where corrosion resistance is crucial.
- \textbf{Automotive:} Applied in high-performance engine parts, turbocharger systems, and exhaust systems. }& 
\small{
Both responses are factually correct but the response generated using RAG is more detailed and this can be explained by the nature of the retrieval mechanism itself.}\\
\hline
\end{tabularx}
\end{singlespace}

\newpage

\begin{singlespace}
\addtocounter{table}{-1}  
\captionof{table}{(continued)} 
\noindent\rule{\textwidth}{0.4pt} 
\vspace{3pt}
\raggedright
\textbf{Query:} What are the different types of lasers used in laser powder bed fusion?\\
\vspace{3pt}
\begin{tabularx}{\linewidth}{@{}>{\raggedright\arraybackslash}p{0.4\linewidth}X>{\raggedright\arraybackslash}p{0.15\linewidth}X>{\raggedright\arraybackslash}p{0.4\linewidth}X@{}}

\hline
\textbf{LLAMA2-7B RAG} & \textbf{GPT-4} & \textbf{Comments} \\ 

\hline
\small{
Based on the information provided, there are several types of lasers used in laser powder bed fusion (LPBF):

\textbf{CO2 lasers:} These are the most common type of lasers used in LPBF. They have a wavelength of 10.6 μm and are suitable for processing metals such as titanium, aluminum, and steel.

\textbf{YAG lasers:} Yttrium aluminum garnet (YAG) lasers have a wavelength of 1.06 μm and are commonly used for processing stainless steel, titanium, and other metals.

\textbf{Nd:YAG lasers:} Neodymium-doped yttrium aluminum garnet (Nd:YAG) lasers have a wavelength of 1.06 μm and are used for processing metals such as stainless steel, titanium, and aluminum.

\textbf{Fiber lasers:} Fiber lasers have a wavelength of 1.06 μm and are used for processing metals such as stainless steel, titanium, and aluminum.

\textbf{Diode lasers:} Diode lasers have a wavelength of 808 nm and are used for processing metals such as aluminum and copper.

\textbf{Hybrid lasers:} Hybrid lasers combine two or more laser sources, such as a CO2 laser and a YAG laser, to create a more powerful and versatile laser system.

The choice of laser type depends on the specific application and the desired properties of the final product. For example, CO2 lasers are commonly used for processing large parts, while YAG lasers are better suited for smaller parts with complex geometries.
} &
\small{
Laser powder bed fusion (LPBF) is a prominent additive manufacturing technology that uses lasers to selectively melt and fuse metallic powder layers. Here are the main types of lasers used in LPBF:

\textbf{$CO_{2}$ Lasers:} These lasers emit at a wavelength of 10.6 micrometers. They are historically significant in early LPBF systems but are less common today due to their larger spot sizes and lower absorption rates by metal powders, which can lead to inefficiencies in energy use and part quality.

\textbf{Fiber Lasers:} Most prevalent in modern LPBF systems, fiber lasers produce a laser beam at a wavelength of about 1.06 micrometers, which is highly absorbed by metals. This type of laser offers high efficiency, a fine focusable beam spot, and high power, enabling faster printing speeds and finer details.

\textbf{Nd:YAG Lasers:} Operating at wavelengths around 1.06 micrometers, similar to fiber lasers, Nd:YAG lasers offer good absorption characteristics for metals. They are less common in the newest machines but are still used for their high peak power and pulse capabilities, useful in certain niche applications.

Each laser type has distinct advantages and applications, influencing their adoption based on specific manufacturing requirements and the properties of the metal powders used.
} &
\small{
The RAG system provides a more detailed response than GPT-4. However, GPT-4 seems more up-to-date about the use of CO2 lasers. GPT-4 is unaware of the application of diode lasers in LPBF.
} \\
\hline

\end{tabularx}
\end{singlespace}

\newpage
\begin{singlespace}
\addtocounter{table}{-1}  
\captionof{table}{(continued)} 
\noindent\rule{\textwidth}{0.4pt} 
\raggedright
\vspace{3pt}
\textbf{Query:} What is the difference between tapped and bulk/apparent density of a powder?\\
\vspace{3pt}
\begin{tabularx}{\linewidth}{@{}>{\raggedright\arraybackslash}p{0.43\linewidth}X>{\raggedright\arraybackslash}p{0.15\linewidth}X>{\raggedright\arraybackslash}p{0.4\linewidth}X@{}}

\hline
\textbf{LLAMA2-7B RAG} & \textbf{GPT-4} & \textbf{Comments} \\ 

\hline
\small{
\textbf{Tapped density refers to the density
of a powder measured after it has
been tapped or agitated}, which can
help to redistribute the particles and
reduce any clumping or
agglomeration. This is important
because the density of a powder can
be affected by the way it is handled
and stored, and tapping or agitating
the powder can help to ensure that
it is uniform and consistent.
\textbf{Bulk density, on the other hand,
refers to the density of a powder
measured in its bulk form, without
any tapping or agitation.} This is the
density of the powder as it is stored
or used in a process, and it can be
affected by factors such as the
particle size distribution, the shape
and size of the particles, and the
surface chemistry of the powder.
In terms of standards, \textbf{ASTM
F3049-14 }can be used to measure
the tapped density of a powder,
while \textbf{ASTM E1317-16 }can be used
to measure the bulk density of a
powder. These standards provide
guidelines for how to measure the
density of a powder using different
techniques, such as the tapped
method or the bulk method.
In terms of the advantages and
disadvantages of each method, the
tapped method can provide more
accurate results for powders with a
narrow particle size distribution,
while the bulk method can provide
more accurate results for powders
with a wider particle size
distribution. The tapped method
can also help to reduce the risk of
contamination or
cross-contamination, while the bulk
method can be faster and more
cost-effective.}
 & 
 \small{
 Tapped density and bulk density both measure the space a powder occupies but differ in how they consider the powder's packing behavior and air voids.

\textbf{Bulk density, or apparent density, is the mass of a powder divided by its total volume}, including the space between particles. This measurement is taken without any external force. To measure bulk density, a known mass of powder is gently poured into a graduated cylinder, and the volume it occupies is measured. This method involves only natural settling, and the resulting bulk density is lower when the particles have irregular shapes or sizes.

\textbf{Tapped density, on the other hand, is measured under conditions where the powder sample is subjected to tapping or vibration} after being poured into a measuring cylinder. The tapping causes the powder particles to settle and reorient themselves to minimize their volume. The tapped density is calculated by dividing the mass of the powder by the settled volume after tapping, which is higher than bulk density as the tapping reduces void spaces.

The difference between bulk and tapped densities helps characterize the flow properties and packing behavior of powders. A high ratio of tapped to bulk density often indicates poor flow properties, while a low ratio suggests better flowability.} & 
\small{
Both responses provide accurate details regarding the differences, but there are more specific details obtained from the RAG system.}\\
\hline
\end{tabularx}
\end{singlespace}

\newpage
\begin{singlespace}
\addtocounter{table}{-1}  
\captionof{table}{(continued)} 
\noindent\rule{\textwidth}{0.4pt} 
\vspace{3pt}
\raggedright
\textbf{Query:} What factors affect creep performance in laser powder bed fusion?\\
\begin{tabularx}{\linewidth}{@{}>{\raggedright\arraybackslash}p{0.44\linewidth}X>{\raggedright\arraybackslash}p{0.12\linewidth}X>{\raggedright\arraybackslash}p{0.44\linewidth}X@{}}

\hline
\textbf{LLAMA2-7B RAG} & \textbf{GPT-4} & \textbf{Comments} \\ 

\hline

\small{
Based on the new context provided, the factors that affect creep performance in laser powder bed fusion (LPBF) can be further refined as:

\textbf{Laser power: }The laser power used in LPBF can affect the creep performance of the material. Higher laser powers can lead to faster creep, while lower laser powers can result in slower creep.
\textbf{Scan speed: }The scan speed of the laser beam over the build platform can also impact creep performance. Faster scan speeds can result in higher creep, while slower scan speeds can lead to lower creep.
\textbf{Temperature: }The temperature of the build platform and the surrounding environment can affect the creep performance of the material. Higher temperatures can lead to increased creep, while lower temperatures can result in slower creep.
\textbf{Material properties: }The properties of the material being LPBF'd, such as its composition and microstructure, can also impact creep performance. For example, materials with a higher density and smaller grain size may exhibit better creep resistance than materials with a lower density and larger grain size.
Laser beam profile: The \textbf{shape and size of the laser beam }used in LPBF can also affect creep performance. A more focused laser beam can lead to higher creep, while a less focused beam can result in slower creep.

Based on the provided references, it is evident that the creep performance of LPBF materials can be affected by a combination of these factors. Further research and experimental studies are needed to fully understand the relationship between these factors and creep performance in LPBF. }& 
\small{
Creep performance in laser powder bed fusion (LPBF), a popular additive manufacturing (AM) method, is influenced by various factors:

\textbf{Material Selection: }The type of metal powder used, such as titanium, nickel, or steel alloys, fundamentally affects creep resistance. Each material has different responses to stress and temperature, impacting its long-term deformation characteristics.
\textbf{Microstructure Control: }LPBF allows precise control over microstructure. The size, orientation, and distribution of grains within the printed object can significantly influence its creep properties. A finer microstructure generally enhances creep resistance.
\textbf{Process Parameters:} Laser power, scanning speed, layer thickness, and hatch spacing affect the thermal gradients and cooling rates experienced during printing. These, in turn, influence the residual stress and microstructural features like porosity and phase composition, all crucial for creep behavior.
\textbf{Post-processing Treatments:} Heat treatments such as annealing or stress-relief can reduce residual stresses and modify the microstructure, thereby improving creep resistance.
\textbf{Design and Geometry: }The design of the component, including its complexity and the orientation of features relative to stress directions, also impacts how it will perform under creep conditions.

Understanding and optimizing these factors is essential for enhancing the creep performance of components manufactured using LPBF.} & 
\small{
Both responses are comparable and satisfactory.
}\\
\hline
\end{tabularx}
\end{singlespace}

\newpage
\begin{singlespace}
\addtocounter{table}{-1}  
\captionof{table}{(continued)} 
\noindent\rule{\textwidth}{0.4pt} 
\vspace{3pt}
\raggedright
\textbf{Query:} What are 5 different methods to produce metal powders? \\
\begin{tabularx}{\linewidth}{@{}>{\raggedright\arraybackslash}p{0.34\linewidth}X>{\raggedright\arraybackslash}p{0.12\linewidth}X>{\raggedright\arraybackslash}p{0.54\linewidth}X@{}}
\hline
\textbf{LLAMA2-7B RAG} & \textbf{GPT-4} & \textbf{Comments} \\ 
\hline
\small{
Based on the provided context information, here are 5 different methods to produce metal powders:
\textbf{Atomization: }This is a common method used to produce metal powders. It involves melting the metal in a furnace and then atomizing it using a nozzle to create small droplets. The droplets are then collected and processed to produce the desired powder.
\textbf{Gas atomization: }This is a variation of the atomization process that uses a gas, such as nitrogen or helium, to atomize the molten metal. This method produces smaller and more uniform powder particles than traditional atomization.
\textbf{Water atomization: }This method involves melting the metal in a furnace and then atomizing it using water to create small droplets. The droplets are then collected and processed to produce the desired powder.
\textbf{Plasma atomization:} This method involves using a plasma arc to melt and atomize the metal. The plasma arc creates a high-temperature plasma that can melt and vaporize the metal, creating small powder particles.
\textcolor{red}{Electrochemical atomization: }This method involves using an electrochemical arc to melt and atomize the metal. The electrochemical arc creates small droplets that are then collected and processed to produce the desired powder.} & 
\small{
Producing metal powders for various industrial applications, including additive manufacturing and metallurgy, involves several techniques, each suited for specific metals and desired powder characteristics. Here are five common methods:
\textbf{Atomization: }This is the most widely used method for producing metal powders. It involves melting the metal and then dispersing it into droplets by a high-speed stream of gas or water. As the droplets cool, they solidify into powders. This method is suitable for a wide range of metals and allows for good control over particle size and shape.
\textbf{Electrolysis: }In this method, metals are deposited from a solution of their salts using an electrical current. The metal deposits onto a cathode, from where it is later removed and crushed into powder. This technique is often used for producing high-purity copper and iron powders.
\textbf{Chemical Reduction: }Powders of metals like iron, tungsten, and molybdenum are often produced by chemical reduction, where a chemical reducing agent reduces metal oxides to metallic powders at high temperatures.
\textbf{Mechanical Alloying: }This involves repeatedly welding, fracturing, and re-welding powders under high-energy ball milling. The process is used for creating alloy powders with controlled microstructures.
\textcolor{red}{Vapor Deposition:} Metal vapors, generated in a vacuum, are condensed onto a cool surface and then collected as powder. This method is particularly useful for producing high-purity metals or special alloys that are difficult to produce through other methods.
Each method offers unique benefits and is chosen based on the specific properties and purity required in the final product.} &
\small{
Both responses are fairly accurate. But some terminology used here is wrong. "Electrochemical atomization" instead of electrode induction melting gas atomization and "Vapor deposition" instead of Vapor phase condensation. 
}\\
\hline
\end{tabularx}
\end{singlespace}

\newpage
\begin{singlespace}
\addtocounter{table}{-1}  
\captionof{table}{(continued)} 
\noindent\rule{\textwidth}{0.4pt} 
\vspace{3pt}
\raggedright
\textbf{Query:} Why is additive manufacturing a feasible production method to generate functionally graded materials?\\
\vspace{3pt}
\begin{tabularx}{\linewidth}{@{}>{\raggedright\arraybackslash}p{0.4\linewidth}X>{\raggedright\arraybackslash}p{0.15\linewidth}X>{\raggedright\arraybackslash}p{0.4\linewidth}X@{}}
\hline
\textbf{LLAMA2-7B RAG} & \textbf{GPT-4} & \textbf{Comments} \\ 
\hline
\small{
The advantages of using AM to produce FGMs include:
\textbf{Complex geometries:} AM allows for the creation of complex geometries that cannot be produced using traditional manufacturing methods. This is particularly useful for FGMs, which often have complex microstructures that require specific geometries to function optimally.
\textbf{Customized material properties: }AM enables the creation of customized material properties by adjusting the composition and microstructure of the material layer by layer. This allows for the creation of FGMs with tailored material properties for specific applications.
\textbf{Precision and accuracy: }AM offers high precision and accuracy, which is essential for creating FGMs with consistent and uniform composition and microstructure.
Reduced material waste: AM produces minimal waste, which reduces the environmental impact of the production process.
\textbf{Rapid prototyping:} AM enables rapid prototyping, which is essential for quickly and efficiently creating FGMs with specific properties for specific applications.
In summary, AM is a feasible production method to generate functionally graded materials with varying compositions and microstructures along specific directions. This is particularly useful in the aerospace and automotive industries, where FGMs are increasingly being used to optimize material properties for specific applications.
} &
\small{
\textbf{ Layer-by-layer fabrication:} AM builds parts layer-by-layer, allowing precise control over material composition and structure at each layer. This capability is critical for creating the gradual transitions in material properties that define FGMs.
\textbf{Material versatility: }AM processes can handle a wide range of materials, including polymers, metals, ceramics, and composites. This flexibility is essential for fabricating FGMs, as it allows for the combination of different materials with distinct physical and mechanical properties within a single component.
\textbf{Design flexibility: A}M supports complex geometries that would be difficult or impossible to achieve with traditional manufacturing methods. This design freedom is beneficial for developing FGMs tailored to specific functional requirements, such as varying porosity or mechanical strength.
\textbf{Digital control: }The digital nature of AM facilitates the integration of computer-aided design (CAD) with the manufacturing process, enabling precise control over the material gradients. This integration is crucial for achieving the desired material distributions and functional properties in FGMs.
\textbf{Cost-effectiveness for small batches: }AM is cost-effective for small production runs and prototypes. This advantage is particularly valuable in the early stages of FGM development, where customized solutions and experimental designs are often needed.} &
\small{
Both responses are factually accurate and comparable.}\\
\hline
\end{tabularx}
\end{singlespace}
\end{document}